\documentclass[11pt]{article}

\usepackage[preprint]{acl}

\pdfoutput=1

\usepackage{times}
\usepackage{latexsym}
\usepackage[T1]{fontenc}
\usepackage[utf8]{inputenc}
\usepackage{microtype}
\usepackage{inconsolata}

\usepackage{amsmath}
\usepackage{amssymb}
\usepackage{amsfonts}
\usepackage{mathtools}
\usepackage{amsthm}
\usepackage{bbm}
\usepackage{nicefrac}

\usepackage{graphicx}
\usepackage{subcaption}
\usepackage{caption}
\usepackage{stfloats}
\usepackage{xcolor}

\usepackage{booktabs}
\usepackage{multirow}
\usepackage{threeparttable}
\usepackage{colortbl}

\usepackage[most]{tcolorbox}
\usepackage{listings}
\usepackage{enumitem}
\usepackage[capitalize,noabbrev]{cleveref}

\theoremstyle{plain}

\theoremstyle{definition}

\theoremstyle{remark}

\title{VisualFLIP: Do Predictions Depend on Task-Critical Visual Evidence in Multimodal Reasoning?}

\author{
  Didi Zhu \quad Changrui Chen \quad Stefanos Zafeiriou \quad Jiankang Deng \\
  Imperial College London
}

\begin{document}
\maketitle


\begin{abstract}
When a multimodal large language model answers a visual reasoning question correctly, is the prediction actually supported by the task-critical visual evidence? Correct answers can coexist with flawed reasoning, making accuracy alone an incomplete test of grounding.
 We introduce VisualFLIP, a paired benchmark with 1{,}374 images arranged as same-question perturbation pairs across cardinality, attribute, spatial, and logic tasks. Each pair keeps the question fixed but minimally changes the evidence so the gold answer deterministically flips.
 We evaluate 24 MLLMs with pair accuracy, which requires solving both sides of a pair, and Collapse Rate (CR), which measures how often a model that solves at least one side repeats the same non-empty answer for both images. Together, these metrics show that paired correctness and evidence dependence are related but distinct: capable models can still fail to update after task-critical visual changes, and collapse becomes more severe for some models when the edited image follows an earlier answer in a sequential setting. Further details are available on our \href{https://didizhu-judy.github.io/VisualFLIP/}{Project page}.
\end{abstract}

\section{Introduction}

\begin{figure}[!t]
  \centering
  \includegraphics[width=\columnwidth]{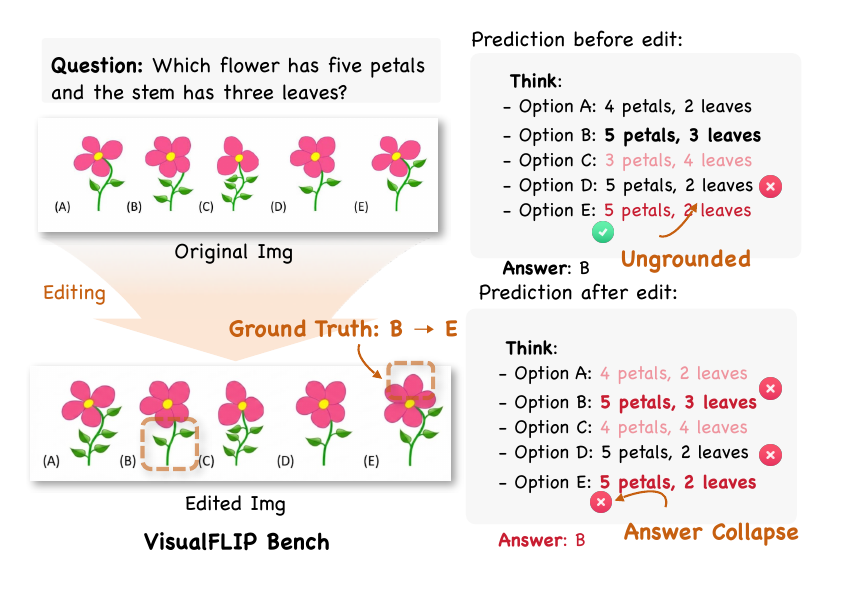}
\caption[Motivating example for VisualFLIP.]{\textbf{Motivating example for VisualFLIP.} The model answers correctly while its reasoning contains counting errors. After a task-critical edit flips the ground truth, the answer stays fixed, revealing an update failure not captured by a single-image correctness check.}
  \label{fig:visual_inertia}
\end{figure}

Modern MLLMs can answer many visual reasoning questions correctly on mathematical, symbolic, and logic-heavy benchmarks \citep{lu2024mathvista,wang2024mathvision,yue2024mmmupro,visulogic2025}; Recent frontier systems also often provide detailed reasoning alongside their final answers \citep{openai2025gpt5,openai2025gpt52,geminiteam2025gemini3,anthropic2025claude45,qwen2025qwen3vl,glm2025glm4v}. Yet a correct answer, even with an apparently detailed reasoning process, does not tell us whether the prediction is supported by the task-critical visual evidence. The answer may be driven by the relevant visual premise; it may also be driven by language priors or memorized benchmark regularities \citep{goyal2017counterfactual,agrawal2018vqacp}. Written reasoning can describe visual content, or even contain local visual mistakes, without showing whether the final answer would change when the evidence it describes changes \citep{verify2025,visulogic2025}.

\begin{figure*}[!t]
  \centering
  \includegraphics[width=0.95\textwidth]{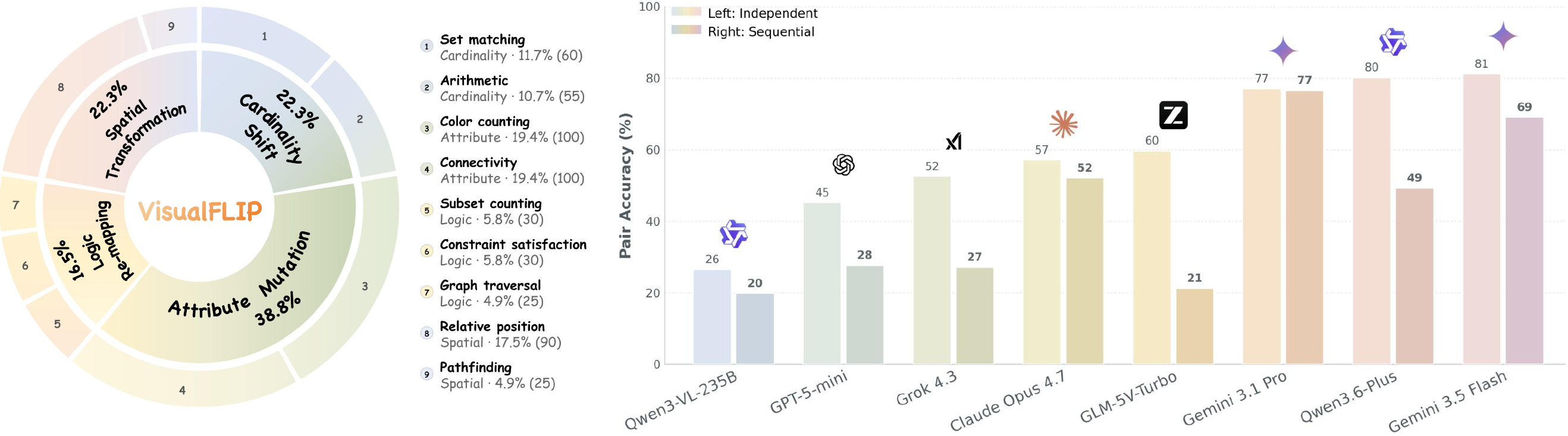}
  \caption[VisualFLIP composition and evaluation modes.]{\textbf{VisualFLIP composition and evaluation modes.} \textit{Left:} sample distribution across four perturbation categories and nine task types. \textit{Right:} pair accuracy for representative MLLMs under independent (light) and sequential (dark) evaluation; prior-answer exposure reduces pair accuracy for several capable systems.}
  \label{fig:vlazy_performance_overview}
\end{figure*}

The question is therefore behavioral. If the visual premise that determines the answer is changed while the question and surrounding context are held fixed, does the model's prediction change? VisualFLIP turns this into an answer-updating test: a prediction that is sensitive to the task-critical evidence should change when that evidence changes.

\textbf{Benchmark.} VisualFLIP organizes image pairs across cardinality, attribute, spatial, and logic perturbations. The benchmark combines source-level synthetic tasks with real-image tasks, so the same paired-flip contract is tested both under controlled rendering and on existing visual-reasoning problems. In each pair, the question is fixed and a minimal task-critical visual change deterministically flips the ground-truth answer, as illustrated in Figure~\ref{fig:visual_inertia}. We report two pair-level measures: pair accuracy (Acc$_\text{p}$), which requires answering both images in a pair correctly, and \textbf{Collapse Rate} (CR), which measures how often a model that answers at least one image correctly repeats the same non-empty answer for both images despite the flipped gold answer in the paired comparison.

\textbf{Findings.} Across 24 MLLMs, including recent frontier systems such as Gemini 3.1 Pro \citep{geminiteam2025gemini3} and Claude Opus 4.7 \citep{anthropic2025claude45}, the paired metrics reveal three patterns: pair accuracy does not determine collapse; sequential exposure can lower pair accuracy and amplify prior-answer persistence; and update failure depends on the changed evidence, with spatial and counting-heavy tasks creating high pressure for visual updating. Tool-augmented models remain vulnerable despite explicit pixel-level operations, and collapse decreases when task-critical edits become more salient.

\textbf{Exploratory mitigation.} We explore Grounded Masking Reinforcement Learning (GMRL) as a training-side mitigation. GMRL masks the task-critical region and uses the KL divergence between original and masked output distributions as a visual-necessity reward, which is intended to encourage sensitivity to edited evidence (Appendix~\ref{sec:gmrl_appendix}).

\textbf{Contributions.}
(1) We formulate answer updating as a paired-perturbation evaluation problem: when the task-critical visual premise changes while the question stays fixed, the model's final answer should change accordingly.
(2) We release VisualFLIP, a paired visual-reasoning benchmark, together with pair accuracy and the Collapse Rate (CR), a symmetric answer-level metric for visual update failure.
(3) We evaluate 24 MLLMs under the independent paired protocol, with a sequential diagnostic, showing that pair accuracy and update failure can diverge across standard and tool-augmented models, especially on spatial and counting-heavy evidence.


\section{Related Work}
\label{sec:related_work}

\paragraph{Visual reasoning benchmarks.} VQA-style benchmarks established image-question answering as a standard diagnostic format \citep{antol2015vqa,goyal2017counterfactual,agrawal2018vqacp,hudson2019gqa}. Recent MLLM suites extend this format to mathematical, symbolic, and logic-heavy visual reasoning \citep{lu2024mathvista,yue2024mmmu,yue2024mmmupro,visulogic2025,fu2024blink}. These datasets are useful for measuring whether a model can answer a given image correctly, but most items are evaluated independently. They therefore do not directly test whether the same question receives a different answer when only the task-critical visual premise changes.

\paragraph{Counterfactual VLM evaluation.} Counterfactual and paired evaluations address this limitation by changing captions, images, objects, or visual premises \citep{kaushik2019learning,thrush2022winoground,zhang2024cvqa,chen2025countervqa,tong2024eyeswideshut,sepehri2025mediconfusion,blinktwice2025}. In VQA-style counterfactual datasets, the usual goal is to break language priors by altering the image or question while preserving a natural QA format. Recent multimodal benchmarks make the intervention more explicit: MVI-Bench injects misleading visual evidence at concept, attribute, and relationship levels to test whether models are distracted by false premises \citep{chen2025mvibench}, while VisualSwap replaces the image during generation to measure whether a model revisits visual evidence after an intervention \citep{shi2026visualswap}. VisualFLIP differs in the unit of measurement: each example is a same-question image pair with a deterministic answer flip, evaluated under an independent paired protocol. This design makes answer updating itself measurable through pair accuracy and the symmetric, competence-conditioned Collapse Rate.

\paragraph{Reasoning traces.} Modern MLLMs often expose reasoning alongside their final answers \citep{openai2025gpt52,geminiteam2025gemini3,anthropic2025claude45,qwen2025qwen3vl,bytedance2025seed15vl}, and recent work studies whether such traces faithfully reflect visual evidence use \citep{verify2025,visulogic2025}. VisualFLIP takes a complementary route: it does not judge the trace itself, but asks whether the final answer changes when the visual evidence that determines the answer changes. This outcome-level test can reveal failures that remain invisible under both single-image accuracy and plausible-looking written reasoning.

\begin{figure*}[!t]
  \centering
  \includegraphics[width=\textwidth]{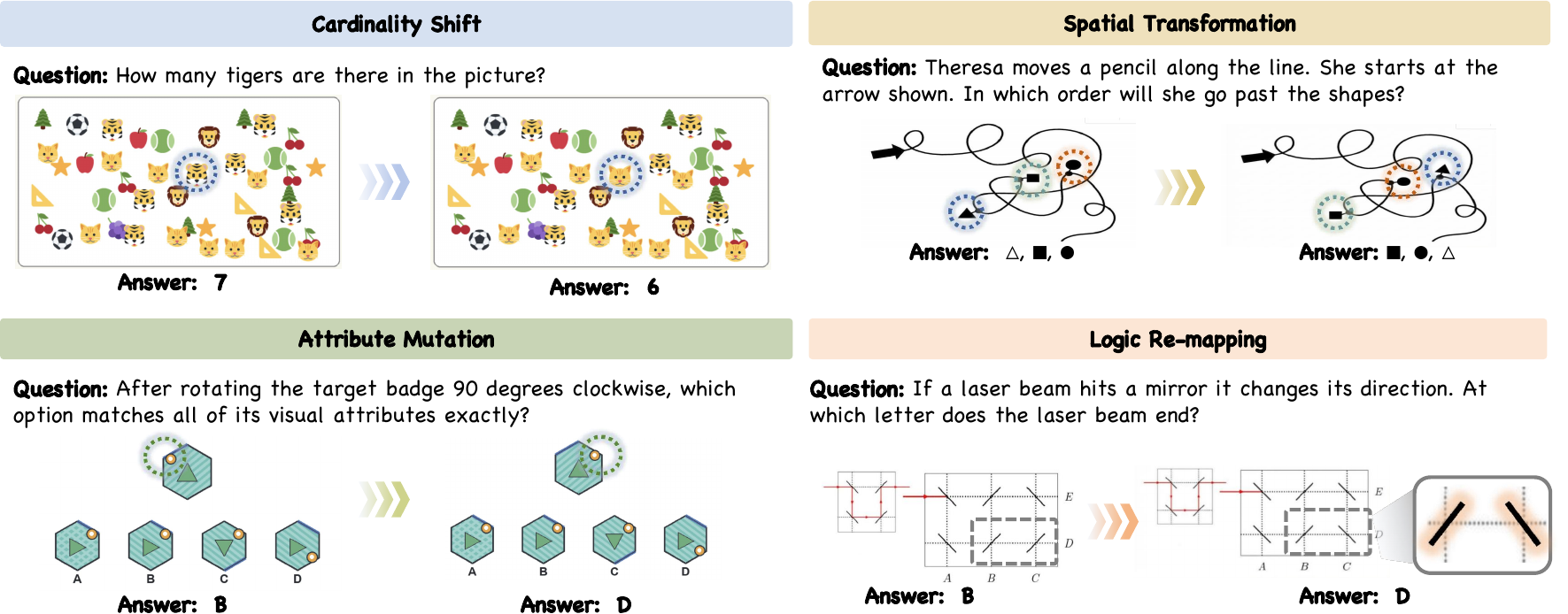}
  \caption[Perturbation types in VisualFLIP.]{\looseness=-1 \textbf{Perturbation Types in VisualFLIP.} VisualFLIP groups perturbations into four categories: \textit{Cardinality Shift} alters object counts; \textit{Spatial Transformation} changes positions or orientations; \textit{Attribute Mutation} modifies visual properties such as color, shape, or value; and \textit{Logic Re-mapping} inverts logical relationships.}
  \label{fig:vlazy_overview}
\end{figure*}

\section{VisualFLIP Benchmark}
\label{sec:benchmark}

VisualFLIP is a collection of minimally perturbed image pairs in which a small visual perturbation deterministically changes the answer. Each pair consists of an original image $v_o$, an edited image $v_e$, a fixed question $q$, and gold answers $(y_o, y_e)$. Because only the task-critical visual premise changes, prediction invariance after the perturbation directly diagnoses failure to use that premise. The benchmark is behavioral; it does not claim to measure mechanistic reasoning.

\subsection{Perturbation Categories}

A useful perturbation must satisfy two opposing constraints. It must be \emph{minimal} in pixel space, so that any answer change is attributable to the perturbed evidence rather than to global distribution shift, residual artifacts, or task-irrelevant context. It must be \emph{maximal} in semantic effect, so that the ground-truth answer flips deterministically and a non-updating prediction can be unambiguously labeled as failure to use the changed evidence. We group such perturbations into four categories, as illustrated in Figure~\ref{fig:vlazy_overview}.

\noindent\textbf{Cardinality Shift.} Changes the number of task-relevant objects (adding, removing, or duplicating instances) while leaving the reasoning template intact, e.g., changing the number of tiger faces in a cluttered scene. These perturbations test whether the model recounts local instances in the updated image rather than carrying over a previous count.

\noindent\textbf{Attribute Mutation.} Changes a premise-bearing visual property, such as a printed value, color, numerical value, or category label, while the surrounding scene stays fixed, e.g., mutating the visual attributes of a target shape so that a different option matches it under rotation. These perturbations test whether the model re-reads the visible attribute from the current image rather than defaulting to a text or category prior.

\noindent\textbf{Spatial Transformation.} Moves or reorients answer-bearing objects, such as trajectory endpoints, stacked arrangements, or an ordering along an axis, e.g., changing the order in which a pencil trajectory passes a set of shapes. The model must recompute the spatial relation from the current image rather than reuse a memorized layout.

\noindent\textbf{Logic Re-mapping.} Changes a visible rule or correspondence, such as a mirror axis, a sorting rule, or a symbol-to-answer key, e.g., reorienting a mirror inside a laser-tracing diagram so the beam endpoint changes. The objects may remain the same, but the rule connecting them to the answer changes, so the model cannot rely on the unchanged objects alone.

Together, the four categories cover the pilot failure axes of counting, label reading, spatial relation tracking, and rule binding. Each category contains both close pairs ($v_o$ and $v_e$ share most pixels) and larger local perturbations; Section~\ref{sec:experiments} reports per-category results so that each contribution can be inspected separately.

\begin{figure*}[!t]
    \centering
    \includegraphics[width=\textwidth]{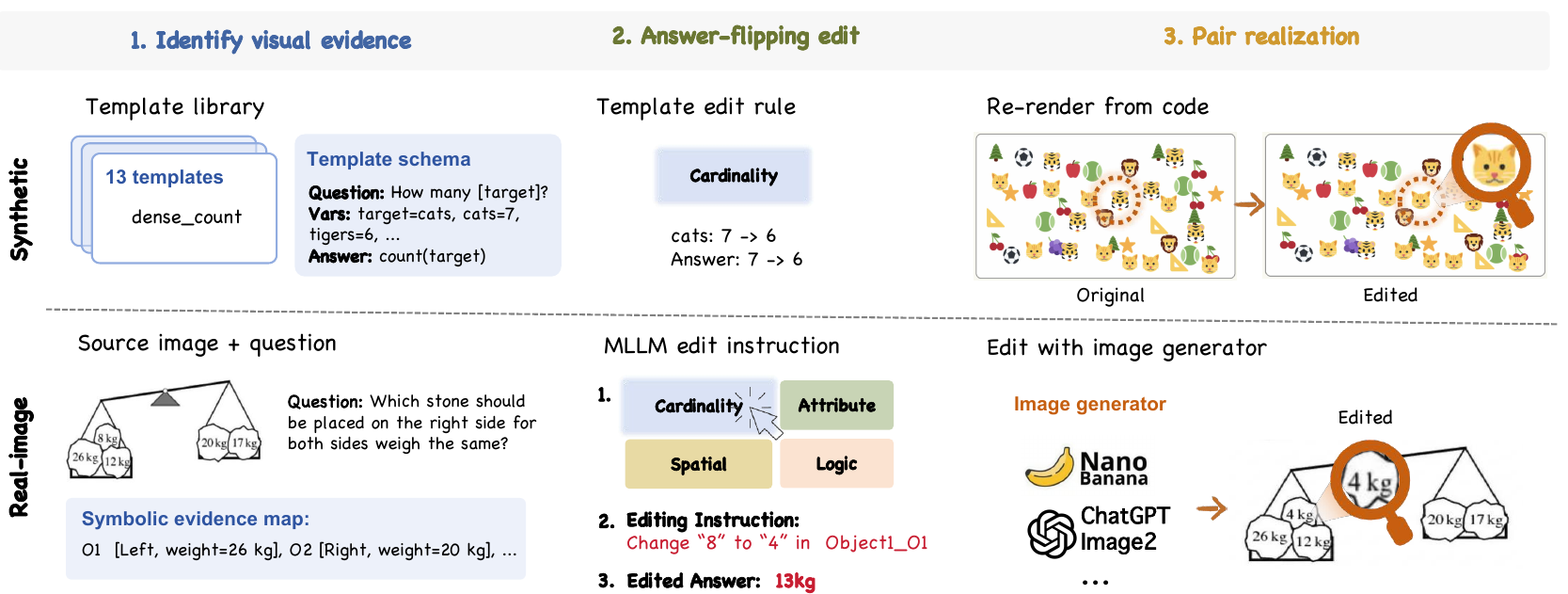}
    \caption[VisualFLIP data construction pipeline.]{\textbf{VisualFLIP Data Construction Pipeline.} Synthetic and real-image sources use different realization routes but share the same paired-flip contract. The released tuple $(v_o, v_e, q, y_o, y_e)$ keeps $q$ fixed, makes the answer flip deterministic, and is manually verified before release.}
    \label{fig:data_generation_pipeline}
\end{figure*}

\definecolor{rowgray}{RGB}{234, 235, 248}
\begin{table*}[!t]
\centering
\renewcommand{\arraystretch}{1.18}
\setlength{\tabcolsep}{5pt}
\scriptsize
\caption[Main results on VisualFLIP.]{\textbf{Main results on VisualFLIP.} We report pair accuracy (Acc$_\text{p}$) and Collapse Rate (CR) across perturbation categories. Higher Acc$_\text{p}$ and lower CR indicate better performance; \textbf{bold} denotes the best result within each group.}
\label{tab:main_results}
\resizebox{\textwidth}{!}{%
\begin{tabular}{lccccccccccc}
\toprule
\multirow{2}{*}{\textbf{Model}} & \multirow{2}{*}{\textbf{Year}} & \multicolumn{2}{c}{\textbf{Cardinality}} & \multicolumn{2}{c}{\textbf{Attribute}} & \multicolumn{2}{c}{\textbf{Spatial}} & \multicolumn{2}{c}{\textbf{Logic}} & \multicolumn{2}{c}{\textbf{Avg}} \\
\cmidrule(lr){3-4} \cmidrule(lr){5-6} \cmidrule(lr){7-8} \cmidrule(lr){9-10} \cmidrule(lr){11-12}
& & {Acc$_\text{p}$}$\uparrow$ & {CR}$\downarrow$ & {Acc$_\text{p}$}$\uparrow$ & {CR}$\downarrow$ & {Acc$_\text{p}$}$\uparrow$ & {CR}$\downarrow$ & {Acc$_\text{p}$}$\uparrow$ & {CR}$\downarrow$ & {Acc$_\text{p}$}$\uparrow$ & {CR}$\downarrow$ \\
\midrule
\multicolumn{12}{c}{\textbf{\textsc{Open-Source}}} \\
\midrule
Qwen2.5-VL-7B & 2025 & 3.4 & 45.2 & 12.8 & 61.2 & 6.7 & 53.6 & 11.0 & 35.2 & 9.2 & 53.0 \\
\rowcolor{rowgray}Qwen3-VL-235B & 2025 & 26.7 & 47.2 & 19.0 & 60.1 & 38.7 & 39.7 & 28.0 & 24.2 & 26.5 & 47.7 \\
Qwen3-VL-32B & 2025 & 27.4 & 50.5 & 16.8 & 67.4 & 42.0 & 36.6 & 19.5 & 37.5 & 25.0 & 51.4 \\
\rowcolor{rowgray}Qwen3-VL-8B & 2025 & 15.8 & 55.1 & 11.7 & 61.5 & 28.0 & 43.2 & 20.3 & 48.5 & 17.6 & 52.7 \\
GLM-4.6V-106B & 2025 & 37.0 & 14.0 & 42.9 & 32.5 & 44.7 & 32.3 & 23.7 & 14.3 & 38.7 & 26.6 \\
\rowcolor{rowgray}Kimi K2.6-1T & 2026 & \textbf{48.6} & \textbf{2.9} & 39.9 & \textbf{1.7} & 41.6 & \textbf{2.4} & 35.6 & \textbf{5.6} & 41.6 & \textbf{2.9} \\
MiMo-v2.5-310B & 2026 & 47.3 & 8.8 & \textbf{59.0} & 7.6 & \textbf{46.2} & 15.7 & \textbf{43.2} & 8.6 & \textbf{51.0} & 9.6 \\
\midrule
\multicolumn{12}{c}{\textbf{\textsc{Open-Source, Tool-Augmented}}} \\
\midrule
\rowcolor{rowgray}DeepEyes-7B & 2025 & 4.1 & 42.1 & 8.1 & 64.6 & 8.0 & 45.6 & 7.6 & 44.3 & 7.1 & 53.0 \\
CoF-7B & 2025 & \textbf{7.5} & \textbf{33.3} & \textbf{13.2} & 54.7 & \textbf{14.7} & 42.0 & 3.4 & 49.0 & \textbf{10.6} & 47.4 \\
\rowcolor{rowgray}PixelReasoner-8B & 2025 & 4.8 & 38.1 & 12.1 & 55.1 & 5.3 & 56.2 & 5.1 & 50.0 & 7.9 & 51.6 \\
Mini-o3-7B & 2025 & 4.8 & 38.3 & 10.6 & \textbf{49.6} & 5.3 & 42.1 & 5.1 & 42.0 & 7.3 & 44.7 \\
\rowcolor{rowgray}DeepEyesV2-7B & 2026 & \textbf{7.5} & 34.4 & 8.4 & 50.7 & 11.3 & \textbf{41.5} & \textbf{10.2} & \textbf{32.8} & 9.2 & \textbf{42.6} \\
\midrule
\multicolumn{12}{c}{\textbf{\textsc{Closed-Source}}} \\
\midrule
GPT-4o & 2024 & 13.0 & 57.1 & 20.9 & 61.7 & 40.0 & 34.7 & 20.3 & 34.9 & 23.3 & 50.6 \\
\rowcolor{rowgray}GPT-5-mini & 2025 & 28.1 & 27.3 & 53.8 & 29.0 & 47.3 & 30.4 & 44.1 & 20.2 & 45.3 & 27.6 \\
Claude Opus 4.6 & 2026 & 19.2 & 25.0 & 38.1 & 17.3 & 30.7 & 29.1 & 14.4 & 32.7 & 28.4 & 23.2 \\
\rowcolor{rowgray}Gemini 3.1 Pro & 2026 & 79.5 & 8.6 & 88.3 & 5.9 & 68.7 & 16.0 & 59.3 & 15.5 & 77.1 & 10.1 \\
Qwen3.5-Flash & 2026 & 74.0 & 6.4 & 87.2 & 3.8 & 52.7 & 17.3 & 59.3 & 7.9 & 72.1 & 7.8 \\
\rowcolor{rowgray}Seed 2.0 Mini & 2026 & 61.7 & 10.1 & 86.2 & 8.8 & 67.4 & 18.4 & 40.2 & 14.5 & 68.6 & 11.8 \\
GLM-5V-Turbo & 2026 & 41.1 & 13.3 & 78.7 & 5.5 & 53.6 & 17.3 & 46.6 & 11.1 & 59.7 & 10.2 \\
\rowcolor{rowgray}Qwen3.6-Plus & 2026 & 83.6 & 8.3 & \textbf{90.5} & \textbf{3.4} & 69.3 & 12.8 & 66.1 & \textbf{5.3} & 80.2 & 6.8 \\
Claude Opus 4.7 & 2026 & 43.2 & 16.0 & 74.0 & 8.3 & 48.7 & 20.3 & 46.6 & 12.2 & 57.2 & 13.1 \\
\rowcolor{rowgray}GPT-5.5 & 2026 & \textbf{85.6} & \textbf{3.7} & 86.4 & 5.7 & 68.0 & \textbf{7.8} & 65.3 & 6.6 & 78.6 & \textbf{5.8} \\
Grok 4.3 & 2026 & 33.6 & 22.9 & 73.6 & 12.0 & 40.0 & 28.8 & 43.2 & 17.2 & 52.5 & 18.4 \\
\rowcolor{rowgray}Gemini 3.5 Flash & 2026 & 84.9 & 7.1 & \textbf{90.5} & 4.5 & \textbf{70.7} & 11.4 & \textbf{68.6} & 9.5 & \textbf{81.2} & 7.3 \\
\bottomrule
\end{tabular}
}
\end{table*}

\subsection{Evaluation Metrics}

We report two pair-level metrics. Pair accuracy, $\text{Acc}_\text{p}$, is the fraction of pairs for which the model answers both images correctly. It is stricter than per-image accuracy because solving one side of a pair does not solve the paired flip.

Collapse Rate (CR) measures update failure among pairs where the model shows at least partial task competence. For a pair $(v_o, v_e)$ with fixed question $q$ and flipped gold answers $y_o \neq y_e$, let $\hat{y}_o$ and $\hat{y}_e$ be the model predictions, and let $c_o,c_e$ indicate correctness on the two sides. A pair is \emph{competent} if $c_o+c_e\geq 1$; among competent pairs, it \emph{collapses} when the model gives the same non-empty answer to both images. We define:
\begin{equation}
\label{eq:collapse_rate}
\mathrm{CR} = \frac{\bigl|\{i : c_o^i + c_e^i \ge 1 \;\wedge\; \hat{y}_o^i=\hat{y}_e^i\neq\emptyset\}\bigr|}{\bigl|\{i : c_o^i + c_e^i \ge 1\}\bigr|}.
\end{equation}
A high CR indicates that the model keeps a single answer across the visual perturbation even though the gold answer changed. Because each VisualFLIP pair is constructed with a deterministic answer flip, the ``original''/``edited'' designation is arbitrary for this metric; CR is invariant to exchanging the two labels, since it conditions only on competence (correct on \emph{either} side) rather than on the original side specifically.

Together, $\text{Acc}_\text{p}$ and CR separate paired correctness from answer updating: $\text{Acc}_\text{p}$ asks whether both images are solved, while CR asks whether a competent model changes its final response when the gold answer flips. We report cross-category and task-type breakdowns in Section~\ref{sec:experiments}; appendix analyses cover fine-grained task and scope details.

\subsection{Data Sources and Construction}

\looseness=-1 VisualFLIP contains two types of pairs: synthetic pairs and real-image pairs. Synthetic pairs are generated from controlled templates, so the visual state and flipped answer are known by construction. Real-image pairs start from existing visual-reasoning images and apply a local perturbation to the task-critical evidence. Both types share the same paired-flip format $(v_o, v_e, q, y_o, y_e)$: the question is fixed, the changed premise flips the gold answer, and the final pair is manually audited.

\noindent\textbf{Step 1: Identify task-critical visual evidence.} For synthetic pairs, construction selects one of 13 hand-authored programmatic templates spanning the four perturbation categories. The selected template instantiates a symbolic state $s_o$ containing the objects, positions, labels, values, relations, distractors, and answer options needed by the question, and provides a renderer $R_t$ and deterministic solver $A_t$, so the original answer is $y_o=A_t(s_o)$. For real-image pairs, the original image and question are first filtered for visual necessity to reduce text-answerable shortcuts; Appendix~\ref{sec:visual_necessity_filter} gives the KL-based criterion. For retained real images, an MLLM identifies the task-relevant objects $\{O_1,\dots,O_n\}$, their locations, and the attributes that the question depends on, producing a symbolic map $\mathcal{S}$ such as ``\texttt{O1: left tray, weight=8kg}''. This map is manually inspected and corrected when needed.

\noindent\textbf{Step 2: Plan an answer-flipping edit.} Given the evidence representation, construction chooses one perturbation category and changes exactly one load-bearing premise while keeping the question fixed. For synthetic examples, this is a template-defined operator $T_c$ applied directly to the symbolic state: \textit{Cardinality Shift} changes a count, \textit{Attribute Mutation} changes a visible value or property, \textit{Spatial Transformation} changes a position or orientation, and \textit{Logic Re-mapping} changes a visible rule or correspondence. The edited state is $s_e=T_c(s_o)$ and the new answer is computed by the same solver, $y_e=A_t(s_e)$. For real-image examples, the same operation is expressed as an edit instruction $\mathcal{E}$ grounded in $\mathcal{S}$, for example changing the value of one object from ``8kg'' to ``4kg'' and deriving the corresponding $y_e$. In both types, the planned perturbation must be minimal in visual scope and must make $y_o \neq y_e$ deterministic rather than merely plausible.

\begin{table*}[!t]
\centering
\renewcommand{\arraystretch}{1.18}
\setlength{\tabcolsep}{5pt}
\scriptsize
\caption{\textbf{Sequential evaluation on VisualFLIP.} Sequential denotes a two-turn protocol in which the model first answers the original image, then receives the perturbed image with the same question in the same conversation. Acc$_\text{p}$ is pair accuracy; SeqCR measures persistence of the original answer after the gold answer flips.}
\label{tab:main_results_seq}
\resizebox{\textwidth}{!}{%
\begin{tabular}{lcccccccccc}
\toprule
\multirow{2}{*}{\textbf{Model}} & \multicolumn{2}{c}{\textbf{Cardinality}} & \multicolumn{2}{c}{\textbf{Attribute}} & \multicolumn{2}{c}{\textbf{Spatial}} & \multicolumn{2}{c}{\textbf{Logic}} & \multicolumn{2}{c}{\textbf{Avg}} \\
\cmidrule(lr){2-3} \cmidrule(lr){4-5} \cmidrule(lr){6-7} \cmidrule(lr){8-9} \cmidrule(lr){10-11}
& {Acc$_\text{p}$}$\uparrow$ & {SeqCR}$\downarrow$ & {Acc$_\text{p}$}$\uparrow$ & {SeqCR}$\downarrow$ & {Acc$_\text{p}$}$\uparrow$ & {SeqCR}$\downarrow$ & {Acc$_\text{p}$}$\uparrow$ & {SeqCR}$\downarrow$ & {Acc$_\text{p}$}$\uparrow$ & {SeqCR}$\downarrow$ \\
\midrule
\rowcolor{rowgray}Qwen3-VL-235B & 11.6 & 72.9 & 13.6 & 68.3 & 36.0 & 39.4 & 24.6 & 38.5 & 19.9 & 56.7 \\
GLM-4.6V & 14.5 & 57.1 & 13.8 & 63.4 & 21.3 & 55.1 & 10.4 & 47.1 & 15.0 & 57.5 \\
\rowcolor{rowgray}GPT-4o & 16.4 & 25.0 & 38.1 & \textbf{8.5} & 45.3 & 11.6 & 14.4 & 15.6 & 31.0 & 13.2 \\
GPT-5-mini & 11.6 & 64.8 & 31.9 & 47.4 & 40.7 & 37.1 & 21.2 & 51.5 & 27.7 & 48.4 \\
\rowcolor{rowgray}Gemini 3.1 Pro & 76.7 & 10.1 & \textbf{83.2} & 10.5 & \textbf{70.7} & 14.5 & \textbf{68.6} & 11.3 & \textbf{76.6} & \textbf{11.4} \\
GLM-5V-Turbo & 18.7 & 79.4 & 19.1 & 38.3 & 28.0 & 41.9 & 21.0 & 33.3 & 21.3 & 47.1 \\
\rowcolor{rowgray}Qwen3.6-Plus & 43.2 & 51.1 & 53.1 & 41.3 & 49.3 & 32.5 & 48.3 & 22.1 & 49.3 & 39.2 \\
Claude Opus 4.7 & 34.9 & 22.5 & 64.5 & 23.4 & 50.0 & 18.6 & 47.5 & 19.8 & 52.1 & 21.6 \\
\rowcolor{rowgray}Grok 4.3 & 15.1 & 41.4 & 26.7 & 43.8 & 40.7 & 27.1 & 26.3 & 25.0 & 27.2 & 36.5 \\
Gemini 3.5 Flash & \textbf{80.8} & \textbf{8.1} & 66.3 & 22.8 & 65.3 & \textbf{9.0} & 66.1 & \textbf{6.8} & 69.1 & 14.2 \\
\bottomrule
\end{tabular}
}
\end{table*}

\noindent\textbf{Step 3: Realize the image pair.} Synthetic pairs are rendered from the two states, $v_o=R_t(s_o)$ and $v_e=R_t(s_e)$, giving source-level control without edit artifacts. Real-image pairs apply $\mathcal{E}$ to the target region of $v_o$ with an image generator and keep the rest of the scene fixed. Both routes yield close pairs with the same question and flipped labels; one renders symbolic states, the other edits an existing image. This shared contract lets us compare failures across construction routes rather than attribute them to the rendering or editing pipeline.

\noindent\textbf{Verification and release.} We store each pair as $(v_o, v_e, q, y_o, y_e)$ and assign a category label. Checks confirm a fixed question, a clear flip, and no other load-bearing change. We drop pairs with edit noise, blur, boundary errors, unclear flips, or ceiling behavior. Cases and prompts appear in App.~\ref{sec:vlaze_examples} and App.~\ref{sec:prompts}.


\section{Experiments}
\label{sec:experiments}

\subsection{Setup}

\textbf{Models.} We evaluate a broad suite of MLLMs, split by \textbf{access} and inference protocol.\footnote{For each model we cite the vendor's model card or technical report for the closest available version; specific point-release cards may not be public at the time of writing.} The \textit{open-source}  group contains standard VLMs such as GLM-4.6V \citep{glm2025glm4v}, Qwen3-VL (235B, 32B, 8B) \citep{qwen2025qwen3vl}, Qwen2.5-VL-7B \citep{qwen2025qwen25vl}, MiMo-v2.5 \citep{xiaomi2025mimovl}, and Kimi K2.6 \citep{moonshot2025kimivl}. The \textit{open-source, tool-augmented} group contains PixelReasoner-8B \citep{su2025pixelreasoner}, DeepEyes-7B \citep{deepeyes2025}, DeepEyesV2-7B \citep{deepeyesv2_2025}, Mini-o3-7B \citep{lai2025minio3}, and CoF-7B \citep{cof2025}; all five are reinforcement-learning fine-tunes of \emph{Qwen2.5-VL-7B-Instruct}, an instruction-tuned model in the Qwen-VL series \citep{bai2023qwenvl,qwen2024vl}, which offers a controlled comparison around a shared backbone. The \textit{closed-source} group includes GPT-4o~\citep{openai2024gpt4o}, GPT-5-mini \citep{openai2025gpt5} and GPT-5.5 \citep{openai2026gpt55}, Claude Opus 4.6 \citep{anthropic2026claude46} and Claude Opus 4.7 \citep{anthropic2026claude47}, Gemini 3.1 Pro \citep{geminiteam2026gemini31pro} and Gemini 3.5 Flash \citep{geminiteam2026gemini35flash}, Grok 4.3 \citep{xai2026grok43}, Qwen3.5-Flash and Qwen3.6-Plus \citep{qwen2025qwen3vl}, GLM-5V-Turbo \citep{glm2025glm4v}, and Seed 2.0 Mini \citep{bytedance2025seed15vl}. Decoding follows a single answer-first protocol with a $4096$-token budget; full details are in Appendix~\ref{sec:eval_impl_app}.

\textbf{Benchmark.} VisualFLIP contains 687 image pairs (1{,}374 images). It includes 515 synthetic pairs from 13 task templates and 172 real-image pairs derived from MathVision. All pairs follow the same answer-flip contract and use four categories: Cardinality, Attribute, Spatial, and Logic. For finer analysis, the synthetic templates are grouped into nine task types in Figures~\ref{fig:vlazy_performance_overview} and~\ref{fig:cr_heatmap}.

\subsection{Main Results on VisualFLIP}

\begin{figure*}[!t]
\centering
\begin{subfigure}[b]{0.54\textwidth}
\centering
\includegraphics[width=\textwidth]{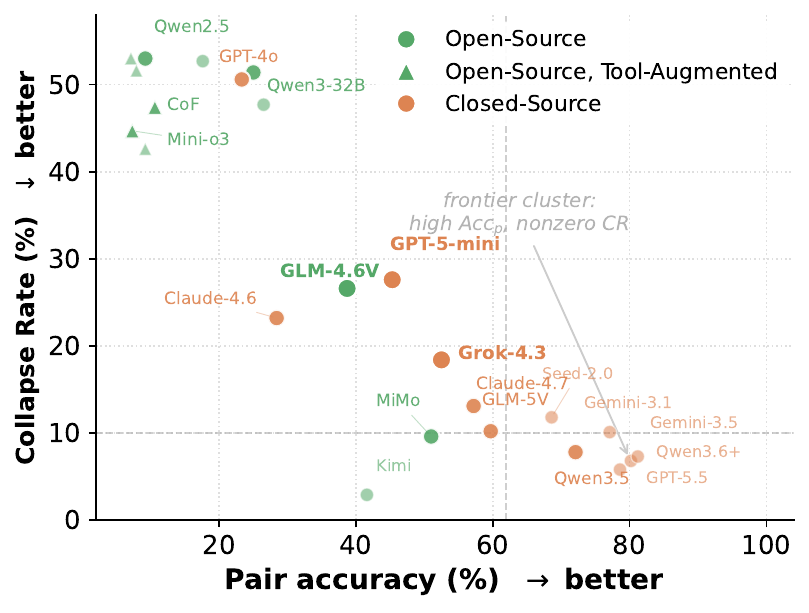}
\caption{Pair accuracy vs.\ CR.}
\label{fig:acc_cr}
\end{subfigure}\hfill
\begin{subfigure}[b]{0.44\textwidth}
\centering
\includegraphics[width=\textwidth]{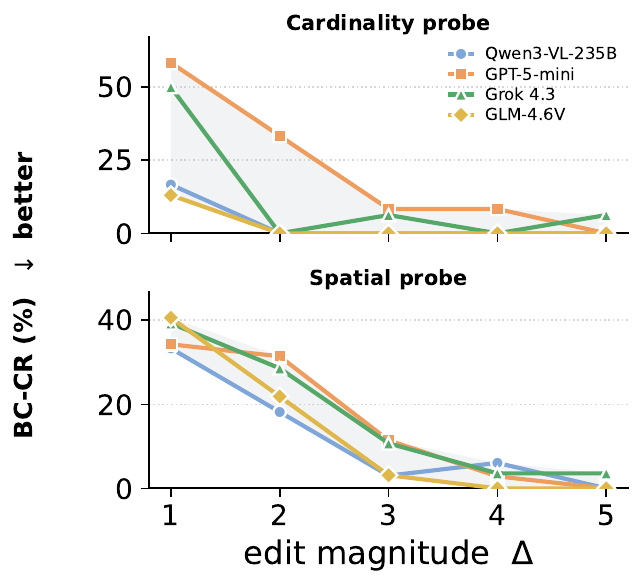}
\caption{Synthetic perturbation-magnitude probes.}
\label{fig:dose}
\end{subfigure}
\caption{\textbf{Pair accuracy vs.\ CR and perturbation magnitude.} \textbf{(a)} Pair accuracy and competence-conditioned CR vary independently across models. \textbf{(b)} Collapse is highest for minimal Cardinality and Spatial perturbations and falls as the changed evidence becomes more salient.}
\label{fig:behavioral}
\end{figure*}

Table~\ref{tab:main_results} summarizes independent-pair evaluation for 24 MLLMs on VisualFLIP. For each perturbation category, we report Acc$_\text{p}$ and CR, with the two images in each pair queried in separate calls.

\textbf{Open-source and tool-augmented models.}
MiMo-v2.5 obtains the highest open-source Acc$_\text{p}$, while Kimi K2.6 achieves the lowest CR, indicating that paired correctness and evidence-dependent updating can improve at different rates. 
The tool-augmented 7B models give a more negative result: relative to their Qwen2.5-VL-7B backbone, they bring only small changes in Acc$_\text{p}$ and CR, and all retain high collapse. 
Thus, in this backbone family, explicit pixel-level operations do not automatically translate into answer decisions that track the changed evidence.

\textbf{Closed-source models.}
Closed-source frontier models form the strongest overall cluster: Gemini 3.5 Flash, Qwen3.6-Plus, and GPT-5.5 all exceed $78\%$ Acc$_\text{p}$ while keeping CR below $8\%$. 
This shows that collapse can be substantially reduced as model capability improves. 
However, the group is not uniform: GPT-5-mini reaches $45.3\%$ Acc$_\text{p}$ but still has $27.6\%$ CR, and Grok 4.3 shows a similar middle regime with $52.5\%$ Acc$_\text{p}$ and $18.4\%$ CR. Thus, CR remains a distinct behavioral signal even among proprietary systems.

\textbf{Category-level trends.}
The category breakdown suggests different failure modes. 
Attribute and Spatial perturbations tend to produce higher CR, consistent with failures in fine-grained visual binding: the model may solve part of the pair, but still repeat an answer when the decisive local evidence changes. 
Cardinality behaves differently: many models solve few complete pairs, so a lower competence-conditioned CR should not be read as the category being easy. 
Instead, counting-heavy examples often expose basic paired-task failure before update failure can even be measured.

\textbf{Outcome decomposition.}
Figure~\ref{fig:outcome_decomp} makes this distinction explicit by decomposing predictions into both-correct, collapse, confused, and both-wrong outcomes. 
This is why Table~\ref{tab:main_results} reports competence-conditioned CR rather than raw same-answer rate: raw rates would mix genuine update failure with both-wrong failures in weak models and both-correct successes in strong models.

\subsection{Analysis of Accuracy and Collapse}

Figure~\ref{fig:acc_cr} summarizes model behavior in the Acc$_\text{p}$ vs. CR plane. The broad trend is expected: the strongest frontier systems occupy the high-accuracy, low-collapse region, while 7B and tool-augmented models cluster at low Acc$_\text{p}$ and high CR. More informative are the deviations from this trend. MiMo-v2.5 and Kimi K2.6 show that paired correctness and collapse do not improve in lockstep: MiMo-v2.5 has the highest open-source Acc$_\text{p}$, whereas Kimi K2.6 has the lowest CR. Conversely, GPT-5-mini and GLM-4.6V sit in a middle regime where nontrivial paired accuracy coexists with substantial collapse. Thus, the scatter is not merely a model-strength ranking; it separates task competence from whether predictions remain controlled by the changed visual evidence.

\subsection{Evidence Salience Analysis}
\label{sec:breakdowns}

\begin{figure}[!t]
\centering
\includegraphics[width=\columnwidth]{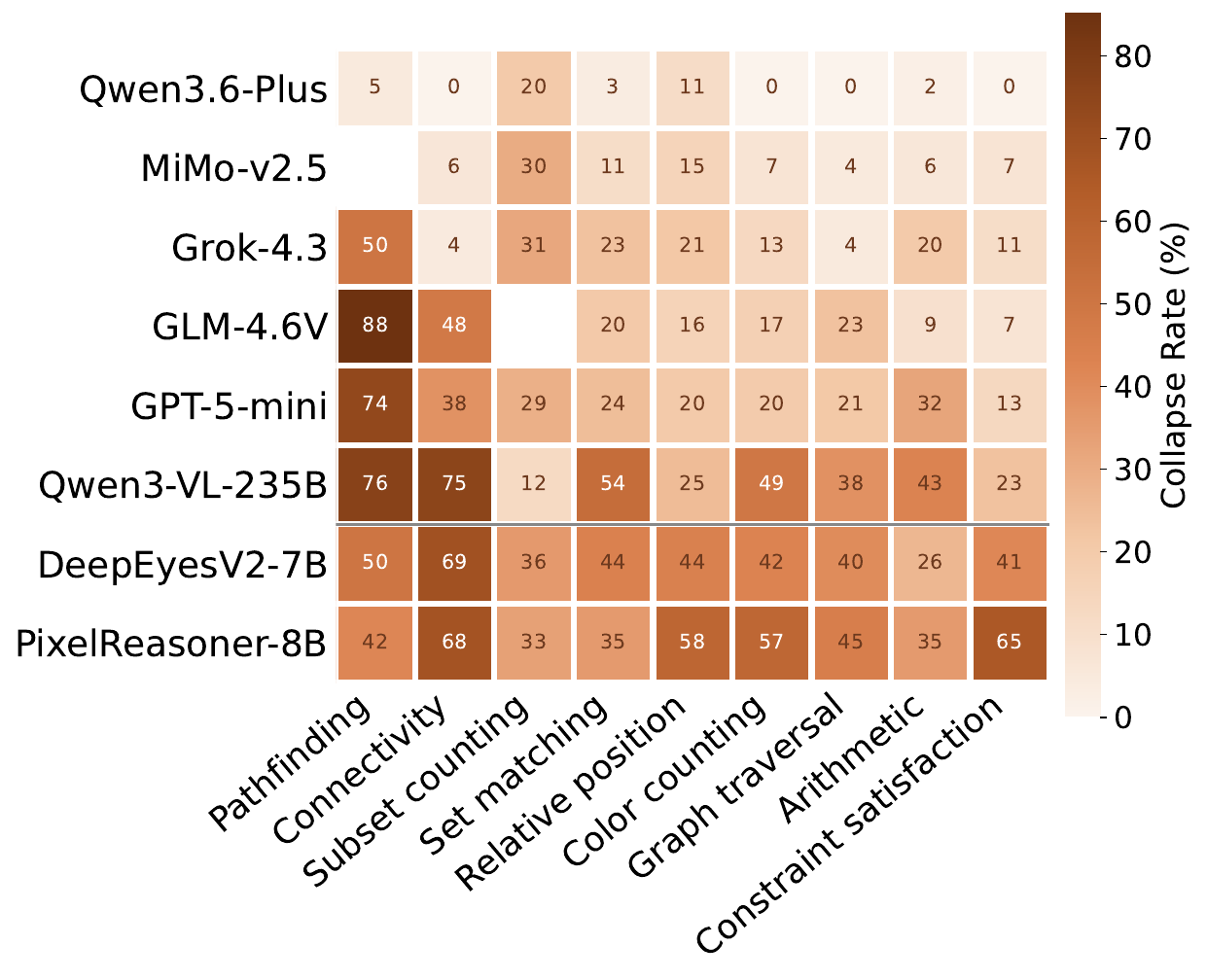}
\caption{\textbf{Task-type CR on the synthetic stratum.} Rows are representative models; columns are task types ordered by mean CR. Darker cells indicate higher CR; blank entries denote $<8$ competent pairs.}
\label{fig:cr_heatmap}

\vspace{0.6em}
\renewcommand{\arraystretch}{1.08}
\setlength{\tabcolsep}{4pt}
\footnotesize
\begin{tabular}{lccc}
\toprule
 & {Task-critical} & \multicolumn{2}{c}{Answer-preserving} \\
\cmidrule(lr){2-2}\cmidrule(lr){3-4}
\textbf{Model} & {Flip$_\text{rel}$}$\uparrow$ & {Keep}$\uparrow$ & {FalseFlip}$\downarrow$ \\
\midrule
\multicolumn{4}{c}{\textbf{\textsc{Open-Source}}} \\
\rowcolor{rowgray} Qwen3-VL-235B & 76 & 80 & 20 \\
GLM-4.6V & 77 & 82 & 18 \\
\midrule
\multicolumn{4}{c}{\textbf{\textsc{Closed-Source}}} \\
\rowcolor{rowgray} GPT-5-mini & 81 & 60 & 40 \\
Grok 4.3 & 76 & 75 & 25 \\
\rowcolor{rowgray} Qwen3.6-Plus & 96 & 98 & 2 \\
\bottomrule
\end{tabular}
\captionof{table}{\textbf{Perturbation locality.} Critical edits flip the gold label; irrelevant edits usually preserve predictions. The Flip$_\text{rel}$--FalseFlip gap ties CR to edited evidence.}
\label{tab:controls}
\vspace{-0.55em}
\end{figure}

To test whether collapse depends on the salience of the visual change, we construct perturbation-magnitude probes on selected synthetic templates. 
For each item, we edit an increasing number of task-critical elements while preserving the question and answer-flip constraint. 
Figure~\ref{fig:dose} shows a dose-response pattern: as more evidence is changed, CR falls in both Cardinality and Spatial settings, suggesting that collapse is most likely when the decisive edit is small or easy to underweight.

Within the synthetic stratum, Figure~\ref{fig:cr_heatmap} further shows that collapse is strongly moderated by task type. 
Pathfinding and connectivity-style tasks are consistently darker, while arithmetic-style tasks are lighter, producing a spread of over $30$ CR points among task types with enough competent pairs. 
Together, the magnitude and task-type analyses suggest that collapse depends on the salience and localizability of the edited evidence, not only on overall visual reasoning accuracy.

\subsection{Perturbation Locality Control}

To test whether CR reflects sensitivity to \emph{task-critical} visual evidence rather than generic answer instability, we compare answer-flipping edits with size-matched perturbations in answer-irrelevant regions (Table~\ref{tab:controls}). Across the evaluated models, predictions change much more often under task-critical perturbations than under irrelevant perturbations. This supports the interpretation that VisualFLIP targets the decisive visual premise rather than arbitrary visual noise. GPT-5-mini remains a useful caveat: its elevated FalseFlip rate indicates nontrivial answer instability, so low CR should still be interpreted together with control behavior.

\subsection{Sequential Evaluation}

The main protocol queries the two images independently, which isolates counterfactual sensitivity without introducing conversational carry-over. We additionally run a sequential diagnostic: the model first answers the original image and then receives the perturbed image with the same question in the same conversation. SeqCR measures whether the second answer repeats the first non-empty answer after the gold answer has flipped.

\looseness=-1 Table~\ref{tab:main_results_seq} shows that sequential context can amplify prior-answer persistence, but the effect is model-specific. The largest increases are for Qwen3.6-Plus, whose SeqCR rises from $6.8\%$ to $39.2\%$, and GPT-5-mini, which rises from $27.6\%$ to $48.4\%$; by contrast, Gemini 3.1 Pro is nearly unchanged, increasing by only $1.3$ points. Table~\ref{tab:seq_tooluse} in the Appendix extends the same protocol to the tool-augmented 7B models, which sit near the paired-accuracy floor, with SeqCR ranging from $76\%$ to $94\%$. Thus, conversational carry-over can introduce an additional failure mode beyond independent-pair collapse, especially for models that strongly anchor on their first answer.


\section{Conclusion}

\looseness=-1 We introduce VisualFLIP, a paired benchmark for testing whether MLLM predictions depend on visual evidence. Across a broad suite of MLLMs, high accuracy and pixel-level tool use do not reliably eliminate collapse: fine-grained attribute and spatial perturbations remain the recurring collapse pressure points, counting-heavy items additionally suppress pair accuracy, and collapse falls as the changed evidence becomes more salient. These results show that high aggregate accuracy is not enough: multimodal reasoning systems must also be tested for whether their predictions depend on the task-critical visual evidence. VisualFLIP therefore complements standard accuracy benchmarks by asking whether a model re-solves the paired instance when the decisive visual premise changes.

\section*{Limitations}

VisualFLIP is a behavioral diagnostic, not a mechanistic test. A high collapse rate identifies failures of evidence-dependent updating, but it does not isolate whether the answer was driven by language priors, memorization, perception errors, or bias from the model-assisted construction and filtering pipeline. Scores for models related to the construction pipeline should therefore be interpreted with this possible overlap in mind. Artifact controls are limited to manual inspection only, without perceptual-divergence metrics or annotator study. Real-image task-type denominators are limited for some evaluated models, and sequential SeqCR is a directional diagnostic with a different denominator from symmetric CR.

\section*{Ethics Statement}

\looseness=-1 VisualFLIP is built from existing visual-reasoning images and model-assisted image edits; we do not collect new images from human subjects or infer sensitive personal attributes. Because benchmark construction uses automated models for filtering, editing, and validation, over-relying on VisualFLIP alone for model selection could favor systems with similar visual priors or editing artifacts. A low Collapse Rate should therefore be interpreted as behavioral robustness on this diagnostic, not as evidence of faithful internal reasoning.

\bibliography{reference}

@inproceedings{bitton2023whoops,
  title={Breaking common sense: Whoops! a vision-and-language benchmark of synthetic and compositional images},
  author={Bitton-Guetta, Nitzan and Bitton, Yonatan and Hessel, Jack and Schmidt, Ludwig and Elovici, Yuval and Stanovsky, Gabriel and Schwartz, Roy},
  booktitle={Proceedings of the IEEE/CVF International Conference on Computer Vision},
  pages={2616--2627},
  year={2023}
}

@article{shao2024deepseekmath,
  title={{DeepSeekMath}: Pushing the Limits of Mathematical Reasoning in Open Language Models},
  author={Shao, Zhihong and Wang, Peiyi and Zhu, Qihao and Xu, Runxin and Song, Junxiao and Bi, Xiao and Zhang, Haowei and Zhang, Mingchuan and Li, Y K and Wu, Y and Guo, Daya},
  journal={arXiv preprint arXiv:2402.03300},
  year={2024}
}

@inproceedings{tishby2015information,
  title={Deep learning and the information bottleneck principle},
  author={Tishby, Naftali and Zaslavsky, Noga},
  booktitle={IEEE Information Theory Workshop (ITW)},
  pages={1--5},
  year={2015}
}

@inproceedings{antol2015vqa,
  title={{VQA}: Visual Question Answering},
  author={Antol, Stanislaw and Agrawal, Aishwarya and Lu, Jiasen and Mitchell, Margaret and Batra, Dhruv and Zitnick, C Lawrence and Parikh, Devi},
  booktitle={International Conference on Computer Vision (ICCV)},
  pages={2425--2433},
  year={2015}
}

@inproceedings{hudson2019gqa,
  title={{GQA}: A New Dataset for Real-World Visual Reasoning and Compositional Question Answering},
  author={Hudson, Drew A and Manning, Christopher D},
  booktitle={Conference on Computer Vision and Pattern Recognition (CVPR)},
  pages={6700--6709},
  year={2019}
}

@article{wang2024mathvision,
  title={Measuring multimodal mathematical reasoning with {Math-Vision} dataset},
  author={Wang, Ke and Pan, Junting and Shi, Weikang and Lu, Zimu and Ren, Houxing and Zhou, Aojun and Zhan, Mingjie and Li, Hongsheng},
  journal={Advances in Neural Information Processing Systems},
  volume={37},
  pages={95095--95169},
  year={2024}
}

@article{qwen2024vl,
  title={{Qwen2-VL}: Enhancing Vision-Language Model's Perception of the World at Any Resolution},
  author={Wang, Peng and Bai, Shuai and Tan, Sinan and Wang, Shijie and Fan, Zhihao and Bai, Jinze and Chen, Keqin and Liu, Xuejing and Wang, Jialin and Ge, Wenbin and Fan, Yang and Dang, Kai and Du, Mengfei and Ren, Xuancheng and Men, Rui and Liu, Dayiheng and Zhou, Chang and Zhou, Jingren and Lin, Junyang},
  journal={arXiv preprint arXiv:2409.12191},
  year={2024}
}

@article{verify2025,
  title={{VERIFY}: A benchmark of visual explanation and reasoning for investigating multimodal reasoning fidelity},
  author={Bi, Jing and Guo, Junjia and Liang, Susan and Sun, Guangyu and Song, Luchuan and Tang, Yunlong and He, Jinxi and Wu, Jiarui and Vosoughi, Ali and Chen, Chen and others},
  journal={arXiv preprint arXiv:2503.11557},
  year={2025}
}

@inproceedings{fu2024blink,
  title={BLINK: Multimodal Large Language Models Can See but Not Perceive},
  author={Fu, Xingyu and others},
  booktitle={arXiv preprint arXiv:2404.12390},
  year={2024},
  note={Foundational work on the perception-reasoning gap}
}

@article{gulati2025putnamaxiom,
  title={{Putnam-AXIOM}: A Functional and Static Benchmark for Measuring Higher Level Mathematical Reasoning in {LLMs}},
  author={Gulati, Aryan and Miranda, Brando and Chen, Eric and Xia, Emily and Fronsdal, Kai and Dumont, Bruno and Obbad, Elyas and Koyejo, Sanmi},
  journal={arXiv preprint arXiv:2508.08292},
  year={2025},
  note={Exposes reliance on memorization through functional variations in university-level mathematics}
}

@inproceedings{lu2024mathvista,
  title={{MathVista}: Evaluating Mathematical Reasoning of Foundation Models in Visual Contexts},
  author={Lu, Pan and Bansal, Hritik and Xia, Tony and Liu, Jiacheng and Li, Chunyuan and Hajishirzi, Hannaneh and Cheng, Hao and Chang, Kai-Wei and Galley, Michel and Gao, Jianfeng},
  booktitle={International Conference on Learning Representations (ICLR)},
  year={2024}
}

@inproceedings{yue2024mmmu,
  title={{MMMU}: A massive multi-discipline multimodal understanding and reasoning benchmark for expert {AGI}},
  author={Yue, Xiang and Ni, Yuansheng and Zhang, Kai and Zheng, Tianyu and Liu, Ruoqi and Zhang, Ge and Stevens, Samuel and Jiang, Dongfu and Ren, Weiming and Sun, Yuxuan and others},
  booktitle={Proceedings of the IEEE/CVF Conference on Computer Vision and Pattern Recognition},
  pages={9556--9567},
  year={2024}
}

@inproceedings{yue2024mmmupro,
  title={{MMMU-Pro}: A more robust multi-discipline multimodal understanding benchmark},
  author={Yue, Xiang and Zheng, Tianyu and Ni, Yuansheng and Wang, Yubo and Zhang, Kai and Tong, Shengbang and Sun, Yuxuan and Yu, Botao and Zhang, Ge and Sun, Huan and others},
  booktitle={Proceedings of the 63rd Annual Meeting of the Association for Computational Linguistics (Volume 1: Long Papers)},
  pages={15134--15186},
  year={2025}
}

@article{chen2025countervqa,
  title={{CounterVQA}: Evaluating and Improving Counterfactual Reasoning in Vision-Language Models for Video Understanding},
  author={Chen, Yuefei and Liu, Jiang and Lin, Xiaodong and Tang, Ruixiang},
  journal={arXiv preprint arXiv:2511.19923},
  year={2025}
}

@inproceedings{zhang2024cvqa,
  title={What if the tv was off? examining counterfactual reasoning abilities of multi-modal language models},
  author={Zhang, Letian and Zhai, Xiaotong and Zhao, Zhongkai and Zong, Yongshuo and Wen, Xin and Zhao, Bingchen},
  booktitle={Proceedings of the IEEE/CVF Conference on Computer Vision and Pattern Recognition},
  pages={21853--21862},
  year={2024}
}

@article{gandhi2024coreknowledge,
  title={Core Knowledge Deficits in Multi-Modal Language Models},
  author={Gandhi, Kanishk and Patel, Raj and Bisk, Yonatan},
  journal={arXiv preprint arXiv:2410.10855},
  year={2024},
  note={Documents failures in counting, perspective-taking, and spatial reasoning}
}

@article{bai2023qwenvl,
  title={{Qwen-VL}: A Versatile Vision-Language Model for Understanding, Localization, Text Reading, and Beyond},
  author={Bai, Jinze and Bai, Shuai and Yang, Shusheng and Wang, Shijie and Tan, Sinan and Wang, Peng and Lin, Junyang and Zhou, Chang and Zhou, Jingren},
  journal={arXiv preprint arXiv:2308.12966},
  year={2023}
}

@article{tang2025chartmuseum,
  title={{ChartMuseum}: Testing Visual Reasoning Capabilities of Large Vision-Language Models},
  author={Tang, Liyan and Kim, Grace and Zhao, Xinyu and Lake, Thom and Ding, Wenxuan and Yin, Fangcong and Singhal, Prasann and Wadhwa, Manya and Liu, Zeyu Leo and Sprague, Zayne and Namuduri, Ramya and Hu, Bodun and Rodriguez, Juan Diego and Peng, Puyuan and Durrett, Greg},
  journal={arXiv preprint arXiv:2505.13444},
  year={2025}
}

@article{visaidmath2024,
  title={{VisAidMath}: Benchmarking visual-aided mathematical reasoning},
  author={Ma, Jingkun and Zhan, Runzhe and Li, Yang and Sun, Di and Chan, Hou Pong and Chao, Lidia S and Wong, Derek F},
  journal={arXiv preprint arXiv:2410.22995},
  year={2024}
}

@inproceedings{agrawal2018vqacp,
  title={Don't Just Assume; Look and Answer: Overcoming Priors in Visual Question Answering},
  author={Agrawal, Aishwarya and Batra, Dhruv and Parikh, Devi and Kembhavi, Aniruddha},
  booktitle={Proceedings of the IEEE Conference on Computer Vision and Pattern Recognition (CVPR)},
  pages={4971----4980},
  year={2018}
}

@inproceedings{goyal2017counterfactual,
  title={Making the V in VQA Matter: Elevating the Role of Image Understanding in Visual Question Answering},
  author={Goyal, Yash and Khot, Tejas and Summers-Stay, Douglas and Batra, Dhruv and Parikh, Devi},
  booktitle={Proceedings of the IEEE Conference on Computer Vision and Pattern Recognition (CVPR)},
  pages={6904----6913},
  year={2017}
}

@inproceedings{kaushik2019learning,
  title={Learning The Difference That Makes A Difference With Counterfactually-Augmented Data},
  author={Kaushik, Divyanshu and Hovy, Eduard and Lipton, Zachary C},
  booktitle={International Conference on Learning Representations (ICLR)},
  year={2020}
}

@inproceedings{thrush2022winoground,
  title={Winoground: Probing Vision and Language Models for Visio-Linguistic Compositionality},
  author={Thrush, Tristan and Jiang, Ryan and Bartolo, Max and Singh, Amanpreet and Williams, Adina and Kiela, Douwe and Rice, Candace},
  booktitle={Proceedings of the IEEE/CVF Conference on Computer Vision and Pattern Recognition (CVPR)},
  pages={5238----5248},
  year={2022}
}

@article{riochet2021intphys,
  title={IntPhys 2019: A Benchmark for Visual Intuitive Physics},
  author={Riochet, Matthieu and Castro, Mario Ynocente and Bernard, Mathieu and Lerer, Adam and Robitaille, Alexandre and Ekonomov, Anton and Grill, Jean-Bastien and Postma, Olivier and Agrawal, Pulkit and Szlam, Arthur and others},
  journal={IEEE Transactions on Pattern Analysis and Machine Intelligence},
  volume={44},
  number={12},
  pages={9201----9215},
  year={2021}
}

@inproceedings{chen2025revisiting,
  title={Revisiting referring expression comprehension evaluation in the era of large multimodal models},
  author={Chen, Jierun and Wei, Fangyun and Zhao, Jinjing and Song, Sizhe and Wu, Bohuai and Peng, Zhuoxuan and Chan, S-H Gary and Zhang, Hongyang},
  booktitle={Proceedings of the Computer Vision and Pattern Recognition Conference},
  pages={513--524},
  year={2025}
}

@article{openai2024gpt4o,
  title={{GPT-4o} system card},
  author={Hurst, Aaron and Lerer, Adam and Goucher, Adam P and Perelman, Adam and Ramesh, Aditya and Clark, Aidan and Ostrow, AJ and Welihinda, Akila and Hayes, Alan and Radford, Alec and others},
  journal={arXiv preprint arXiv:2410.21276},
  year={2024}
}

@article{openai2025gpt5,
  title={{OpenAI GPT-5} System Card},
  author={Singh, Aaditya and Fry, Adam and Perelman, Adam and Tart, Adam and Ganesh, Adi and El-Kishky, Ahmed and McLaughlin, Aidan and Low, Aiden and Ostrow, AJ and Ananthram, Akhila and others},
  journal={arXiv preprint arXiv:2601.03267},
  year={2026}
}

@misc{openai2025gpt52,
  title={Introducing GPT-5.2},
  author={{OpenAI}},
  year={2025},
  url={https://openai.com/index/introducing-gpt-5-2/}
}

@misc{anthropic2025claude45,
  title={Claude Sonnet 4.5 System Card},
  author={{Anthropic}},
  year={2025},
  url={https://www.anthropic.com/claude-sonnet-4-5-system-card}
}

@misc{anthropic2026claude46,
  title={Claude Opus 4.6 System Card},
  author={{Anthropic}},
  year={2026},
  url={https://www.anthropic.com/claude-opus-4-6-system-card}
}

@misc{anthropic2026claude47,
  title={Claude Opus 4.7 System Card},
  author={{Anthropic}},
  year={2026},
  url={https://www.anthropic.com/claude-opus-4-7-system-card}
}

@misc{geminiteam2025gemini3,
  title={Gemini 3 Pro Model Card},
  author={{Gemini Team, Google DeepMind}},
  year={2025},
  url={https://storage.googleapis.com/deepmind-media/Model-Cards/Gemini-3-Pro-Model-Card.pdf}
}

@misc{geminiteam2026gemini31pro,
  title={Gemini 3.1 Pro Model Card},
  author={{Gemini Team, Google DeepMind}},
  year={2026},
  url={https://storage.googleapis.com/deepmind-media/Model-Cards/Gemini-3-1-Pro-Model-Card.pdf}
}

@misc{geminiteam2026gemini35flash,
  title={Gemini 3.5 Flash Model Card},
  author={{Gemini Team, Google DeepMind}},
  year={2026},
  url={https://storage.googleapis.com/deepmind-media/Model-Cards/Gemini-3-5-Flash-Model-Card.pdf}
}

@article{bytedance2025seed15vl,
  title={{Seed1.5-VL} technical report},
  author={Guo, Dong and Wu, Faming and Zhu, Feida and Leng, Fuxing and Shi, Guang and Chen, Haobin and Fan, Haoqi and Wang, Jian and Jiang, Jianyu and Wang, Jiawei and others},
  journal={arXiv preprint arXiv:2505.07062},
  year={2025}
}

@article{moonshot2025kimivl,
  title={{Kimi-VL} Technical Report},
  author={{Kimi Team}},
  journal={arXiv preprint arXiv:2504.07491},
  year={2025}
}

@misc{xai2026grok43,
  title={Grok 4.3 Model Card},
  author={{xAI}},
  year={2026},
  url={https://data.x.ai/grok-4-3-model-card.pdf}
}

@misc{openai2026gpt55,
  title={Introducing GPT-5.5},
  author={{OpenAI}},
  year={2026},
  url={https://openai.com/index/introducing-gpt-5-5/}
}

@article{su2025pixelreasoner,
  title={Pixel Reasoner: Incentivizing Pixel-Space Reasoning with Curiosity-Driven Reinforcement Learning},
  author={Su, Alex and Wang, Haozhe and Ren, Weiming and Lin, Fangzhen and Chen, Wenhu},
  journal={arXiv preprint arXiv:2505.15966},
  year={2025}
}

@article{deepeyes2025,
  title={{DeepEyes}: Incentivizing ``Thinking with Images'' via Reinforcement Learning},
  author={Zheng, Ziwei and Yang, Michael and Hong, Jack and Zhao, Chenxiao and Xu, Guohai and Yang, Le and Shen, Chao and Yu, Xing},
  journal={arXiv preprint arXiv:2505.14362},
  year={2025}
}

@article{deepeyesv2_2025,
  title={{DeepEyesV2}: Toward Agentic Multimodal Model},
  author={Hong, Jack and Zhao, Chenxiao and Zhu, ChengLin and Lu, Weiheng and Xu, Guohai and Yu, Xing},
  journal={arXiv preprint arXiv:2511.05271},
  year={2025}
}

@article{lai2025minio3,
  title={{Mini-O3}: Scaling Up Reasoning Patterns and Interaction Turns for Visual Search},
  author={Lai, Xin and Li, Junyi and Li, Wei and Liu, Tao and Li, Tianjian and Zhao, Hengshuang},
  journal={arXiv preprint arXiv:2509.07969},
  year={2025}
}

@article{cof2025,
  title={Chain-of-Focus: Adaptive Visual Search and Zooming for Multimodal Reasoning via RL},
  author={Zhang, Xintong and Gao, Zhi and Zhang, Bofei and Li, Pengxiang and Zhang, Xiaowen and Liu, Yang and Yuan, Tao and Wu, Yuwei and Jia, Yunde and Zhu, Song-Chun and others},
  journal={arXiv preprint arXiv:2505.15436},
  year={2025}
}

@misc{glm2025glm4v,
      title={{GLM-4.5V} and {GLM-4.1V-Thinking}: Towards Versatile Multimodal Reasoning with Scalable Reinforcement Learning},
      author={{GLM-V Team}},
      year={2025},
      eprint={2507.01006},
      archivePrefix={arXiv},
      primaryClass={cs.CV},
      url={https://arxiv.org/abs/2507.01006},
}

@article{qwen2025qwen25vl,
  title={{Qwen2.5-VL} technical report},
  author={Bai, Shuai and Chen, Keqin and Liu, Xuejing and Wang, Jialin and Ge, Wenbin and Song, Sibo and Dang, Kai and Wang, Peng and Wang, Shijie and Tang, Jun and others},
  journal={arXiv preprint arXiv:2502.13923},
  year={2025}
}

@article{qwen2025qwen3vl,
  title={{Qwen3-VL} Technical Report},
  author={{Qwen Team}},
  journal={arXiv preprint arXiv:2511.21631},
  year={2025}
}

@inproceedings{wu2024vstar,
  title={V*: Guided Visual Search as a Core Mechanism in Multimodal LLMs},
  author={Wu, Penghao and Xie, Saining},
  booktitle={Conference on Computer Vision and Pattern Recognition (CVPR)},
  year={2024}
}

@article{blinktwice2025,
  title={{BLINK-Twice}: You see, but do you observe? A Reasoning Benchmark on Visual Perception},
  author={Ye, Junyan and Jiang, Dongzhi and He, Jun and Zhou, Baichuan and Huang, Zilong and Yan, Zhiyuan and Li, Hongsheng and He, Conghui and Li, Weijia},
  journal={arXiv preprint arXiv:2510.09361},
  year={2025}
}

@article{visulogic2025,
  title={{VisuLogic}: A benchmark for evaluating visual reasoning in multi-modal large language models},
  author={Xu, Weiye and Wang, Jiahao and Wang, Weiyun and Chen, Zhe and Zhou, Wengang and Yang, Aijun and Lu, Lewei and Li, Houqiang and Wang, Xiaohua and Zhu, Xizhou and others},
  journal={arXiv preprint arXiv:2504.15279},
  year={2025}
}

@article{muslr2025,
  title={MuSLR: Multimodal Symbolic Logical Reasoning},
  author={Xu, Jundong and Fei, Hao and Zhang, Yuhui and Pan, Liangming and Huang, Qijun and Liu, Qian and Nakov, Preslav and Kan, Min-Yen and Wang, William Yang and Lee, Mong-Li and others},
  journal={arXiv preprint arXiv:2509.25851},
  year={2025}
}

@inproceedings{tong2024eyeswideshut,
  title={Eyes Wide Shut? Exploring the Visual Shortcomings of Multimodal LLMs},
  author={Tong, Shengbang and Liu, Zhuang and Zhai, Yuexiang and Ma, Yi and LeCun, Yann and Xie, Saining},
  booktitle={Proceedings of the IEEE/CVF Conference on Computer Vision and Pattern Recognition},
  pages={9568--9578},
  year={2024}
}

@inproceedings{sepehri2025mediconfusion,
  title={MediConfusion: Can You Trust Your AI Radiologist? Probing the Reliability of Multimodal Medical Foundation Models},
  author={Sepehri, Mohammad Shahab and Fabian, Zalan and Soltanolkotabi, Maryam and Soltanolkotabi, Mahdi},
  booktitle={International Conference on Learning Representations},
  year={2025}
}

@misc{xiaomi2025mimovl,
  title={{MiMo-VL} Technical Report},
  author={{Xiaomi LLM-Core Team}},
  year={2025},
  eprint={2506.03569},
  archivePrefix={arXiv},
  primaryClass={cs.CL},
  url={https://arxiv.org/abs/2506.03569}
}

@article{chen2025mvibench,
  title={{MVI-Bench}: A Comprehensive Benchmark for Evaluating Robustness to Misleading Visual Inputs in {LVLMs}},
  author={Chen, Huiyi and Peng, Jiawei and Min, Dehai and Sun, Changchang and Chen, Kaijie and Yan, Yan and Yang, Xu and Cheng, Lu},
  journal={arXiv preprint arXiv:2511.14159},
  year={2025}
}

@article{shi2026visualswap,
  title={Are {VLMs} Seeing or Just Saying? Uncovering the Illusion of Visual Re-examination},
  author={Shi, Chufan and Yang, Cheng and Wu, Yaokang and Jin, Linhao and Shui, Bo and Berg-Kirkpatrick, Taylor and Ma, Xuezhe},
  journal={arXiv preprint arXiv:2605.15864},
  year={2026}
}

\clearpage
\appendix


\section*{Appendix Overview}

This appendix is organized into three blocks: companion analyses for the main results, benchmark and construction details, and the preliminary GMRL mitigation study.

\begin{itemize}[leftmargin=12pt,itemsep=2pt,topsep=2pt]
    \item \textbf{Appendix~\ref{sec:main_result_appendix}} gives companion analyses and evaluation implementation details for the main VisualFLIP results.
    \item \textbf{Appendix~\ref{sec:benchmark_details_app}} positions VisualFLIP relative to existing benchmarks and documents construction materials.
    \item \textbf{Appendix~\ref{sec:gmrl_appendix}} describes the exploratory GMRL mitigation study, with sampling-time analysis.
\end{itemize}

\section{Main Result Companion Analyses}
\label{sec:main_result_appendix}

\subsection{Outcome Decomposition}
\label{sec:outcome_decomp_app}

\begin{figure*}[!b]
\centering
\includegraphics[width=0.82\textwidth]{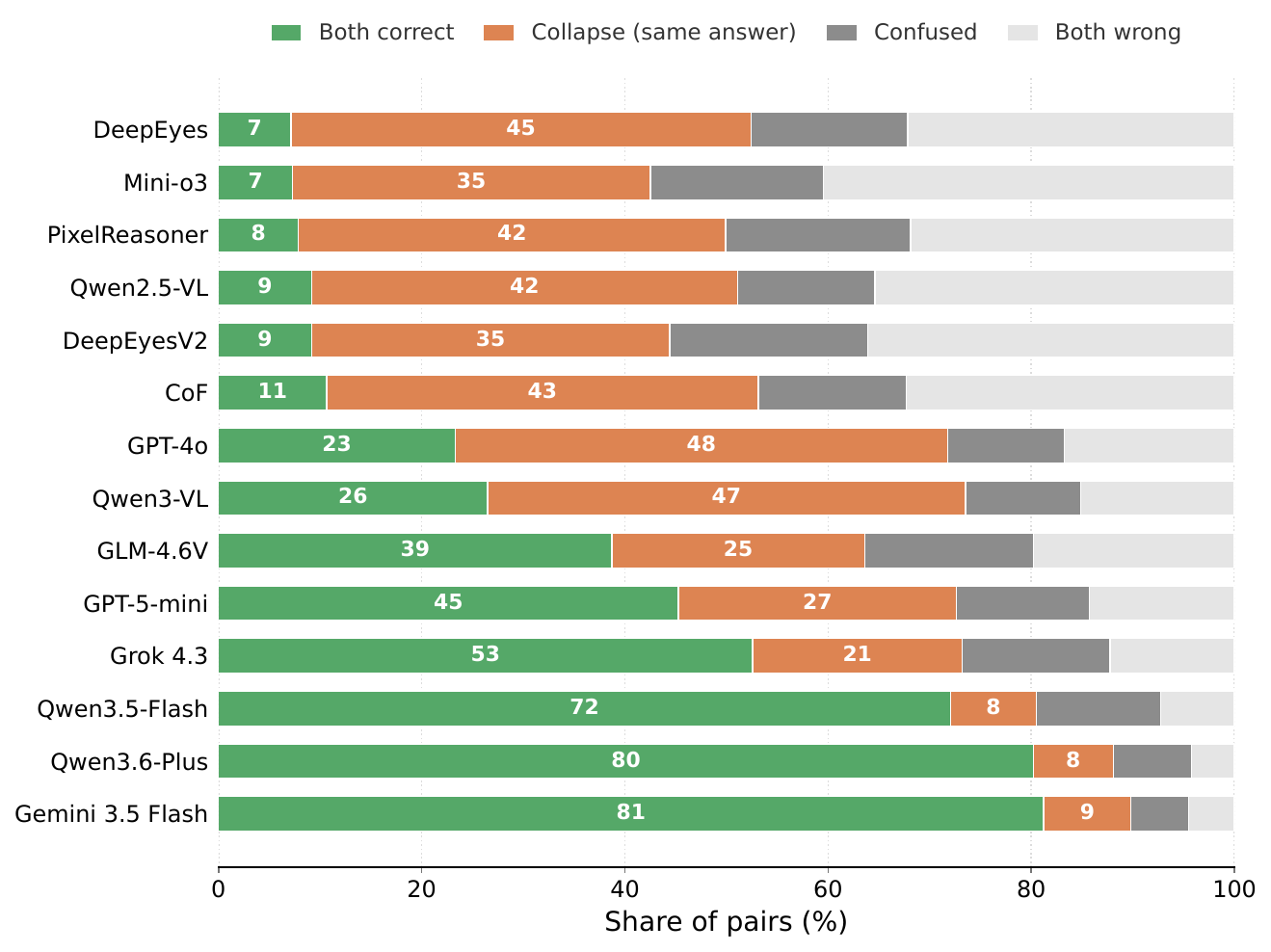}
\caption[Per-pair outcome decomposition.]{\textbf{Per-pair outcome decomposition.} Models with pair-level logs; each pair is one of four outcomes: \emph{both correct}, \emph{collapse} (same answer on both images), \emph{confused} (exactly one side correct), or \emph{both wrong}. The orange segment is an unconditioned same-answer rate over all pairs; Table~\ref{tab:main_results} reports the competence-conditioned Collapse Rate.}
\label{fig:outcome_decomp}
\end{figure*}

Figure~\ref{fig:outcome_decomp} decomposes complete pair-level logs into mutually exclusive outcomes. This diagnostic complements Table~\ref{tab:main_results}: the main table reports competence-conditioned CR, while this figure shows how collapse coexists with both-correct, confused, and both-wrong outcomes over all pairs.

\subsection{Per-Image Accuracy}
\label{sec:per_image_acc}

The main results (Table~\ref{tab:main_results}) report \emph{pair accuracy} (Acc$_\text{p}$), the fraction of counterfactual pairs answered correctly on \emph{both} images, because the unit of evaluation in VisualFLIP is the pair. For comparability with single-image benchmarks, Table~\ref{tab:per_image_acc} reports standard \emph{per-image} accuracy on the same $N{=}687$ evaluation set. We omit CR here to avoid duplicating Table~\ref{tab:main_results}. Per-image accuracy is uniformly higher than pair accuracy (a model can answer one side of a pair correctly without solving the pair), but the two metrics induce essentially the same model ordering.

\begin{table*}[!t]
\centering
\renewcommand{\arraystretch}{1.02}
\setlength{\tabcolsep}{2pt}
\footnotesize
\caption[Per-image accuracy companion.]{\textbf{Per-image accuracy companion.} Main table reports \emph{pair} accuracy (Acc$_\text{p}$: both images of a counterfactual pair correct); here we report \emph{per-image} accuracy Acc\,$=\,(\text{Acc}_\text{o}+\text{Acc}_\text{e})/2$ for comparability with single-image benchmarks. Collapse Rate is not repeated here because it is reported in Table~\ref{tab:main_results}. \textbf{bold} = best per group.}
\label{tab:per_image_acc}
\begin{minipage}{0.90\textwidth}
\centering
\begin{tabular}{@{}>{\raggedright\arraybackslash}p{0.24\linewidth}>{\centering\arraybackslash}p{0.085\linewidth}>{\centering\arraybackslash}p{0.12\linewidth}>{\centering\arraybackslash}p{0.12\linewidth}>{\centering\arraybackslash}p{0.12\linewidth}>{\centering\arraybackslash}p{0.12\linewidth}>{\centering\arraybackslash}p{0.12\linewidth}@{}}
\toprule
\textbf{Model} & \textbf{Year} & \textbf{Card.} & \textbf{Attr.} & \textbf{Spatial} & \textbf{Logic} & \textbf{Avg} \\
\midrule
\multicolumn{7}{c}{\textbf{\textsc{Open-Source}}} \\
\midrule
Qwen2.5-VL-7B & 2025 & 16.1 & 34.2 & 31.3 & 28.4 & 28.7 \\
\rowcolor{rowgray}Qwen3-VL-235B & 2025 & 49.7 & 46.7 & 61.3 & \textbf{40.3} & 49.4 \\
Qwen3-VL-32B & 2025 & 50.3 & 43.8 & \textbf{65.7} & \textbf{40.3} & 49.3 \\
\rowcolor{rowgray}Qwen3-VL-8B & 2025 & 34.6 & 29.7 & 51.0 & 38.1 & 36.8 \\
GLM-4.6V-106B & 2025 & \textbf{52.7} & \textbf{63.7} & 63.7 & 38.6 & \textbf{57.1} \\
\rowcolor{rowgray}Kimi K2.6-1T & 2026 & 60.3 & 54.6 & 54.8 & 48.3 & 54.8 \\
MiMo-v2.5-310B & 2026 & 62.3 & 72.9 & 60.3 & 55.9 & 65.0 \\
\midrule
\multicolumn{7}{c}{\textbf{\textsc{Open-Source, Tool-Augmented}}} \\
\midrule
\rowcolor{rowgray}DeepEyes-7B & 2025 & 21.6 & 30.4 & 26.7 & \textbf{29.7} & 27.6 \\
CoF-7B & 2025 & 21.2 & 33.7 & \textbf{34.3} & 22.5 & 29.3 \\
\rowcolor{rowgray}PixelReasoner-8B & 2025 & 24.0 & \textbf{35.0} & 32.3 & 28.0 & \textbf{30.9} \\
Mini-o3-7B & 2025 & 18.5 & 28.9 & 28.0 & 23.7 & 25.6 \\
\rowcolor{rowgray}DeepEyesV2-7B & 2026 & \textbf{24.7} & 30.2 & 33.0 & \textbf{29.7} & 29.5 \\
\midrule
\multicolumn{7}{c}{\textbf{\textsc{Closed-Source}}} \\
\midrule
GPT-4o & 2024 & 37.7 & 48.7 & 60.3 & 36.9 & 46.9 \\
\rowcolor{rowgray}GPT-5-mini & 2025 & 47.9 & 69.2 & 68.7 & 59.7 & 63.0 \\
Claude Opus 4.6 & 2026 & 37.0 & 59.3 & 52.0 & 30.5 & 48.0 \\
\rowcolor{rowgray}Gemini 3.1 Pro & 2026 & 87.7 & 94.0 & 82.3 & 70.8 & 86.1 \\
Qwen3.5-Flash & 2026 & 84.9 & 92.3 & 70.7 & 67.4 & 81.7 \\
\rowcolor{rowgray}Seed 2.0 Mini & 2026 & 76.2 & 92.6 & 81.0 & 55.7 & 80.0 \\
GLM-5V-Turbo & 2026 & 54.1 & 84.4 & 68.4 & 57.6 & 69.7 \\
\rowcolor{rowgray}Qwen3.6-Plus & 2026 & \textbf{91.1} & 94.3 & 81.7 & 73.3 & 87.3 \\
Claude Opus 4.7 & 2026 & 62.3 & 83.3 & 67.0 & 61.4 & 71.5 \\
\rowcolor{rowgray}GPT-5.5 & 2026 & 89.4 & 91.6 & 76.7 & 71.2 & 84.4 \\
Grok 4.3 & 2026 & 54.1 & 82.8 & 59.3 & 58.5 & 67.4 \\
\rowcolor{rowgray}Gemini 3.5 Flash & 2026 & 90.4 & \textbf{94.5} & \textbf{82.0} & \textbf{74.6} & \textbf{87.5} \\
\bottomrule
\end{tabular}
\end{minipage}
\end{table*}

\subsection{Sequential Evaluation of Tool-Augmented Models}
\label{sec:seq_tooluse}

Table~\ref{tab:main_results_seq} reports sequential evaluation for the branded model suite. Table~\ref{tab:seq_tooluse} extends the same protocol (original then edited image in a single conversation, answer-first decoding, $1024$ tokens, all $N{=}687$ pairs) to the tool-augmented and 7B baseline models. In the paired setting these systems are near the accuracy floor (Acc$_\text{p}<6\%$), and under sequential exposure they exhibit extreme original-answer persistence: SeqCR ranges from $76\%$ to $94\%$, meaning that once they commit to an answer on the original image they almost never revise it after the edit. SeqCR (sequential original-answer persistence) is not identical to the symmetric Collapse Rate of Table~\ref{tab:main_results}, so we report it only as a sequential-setting diagnostic.

\begin{table*}[!t]
\centering
\renewcommand{\arraystretch}{1.18}
\setlength{\tabcolsep}{5pt}
\scriptsize
\caption[Sequential evaluation of tool-augmented models.]{\textbf{Sequential evaluation of tool-augmented and 7B baseline models.} Same protocol as Table~\ref{tab:main_results_seq} (original then edited image in conversation; answer-first, $1024$ tokens; all $N{=}687$ pairs). Acc$_\text{p}$ is pair accuracy; SeqCR is original-answer persistence after edit. All models sit near the paired-accuracy floor and show high persistence (SeqCR $76$--$94\%$): once committed to an answer on the original image, they almost never revise it after the edit.}
\label{tab:seq_tooluse}
\resizebox{\textwidth}{!}{%
\begin{tabular}{lcccccccccc}
\toprule
\multirow{2}{*}{\textbf{Model}} & \multicolumn{2}{c}{\textbf{Cardinality}} & \multicolumn{2}{c}{\textbf{Attribute}} & \multicolumn{2}{c}{\textbf{Spatial}} & \multicolumn{2}{c}{\textbf{Logic}} & \multicolumn{2}{c}{\textbf{Avg}} \\
\cmidrule(lr){2-3} \cmidrule(lr){4-5} \cmidrule(lr){6-7} \cmidrule(lr){8-9} \cmidrule(lr){10-11}
& {Acc$_\text{p}$}$\uparrow$ & {SeqCR}$\downarrow$ & {Acc$_\text{p}$}$\uparrow$ & {SeqCR}$\downarrow$ & {Acc$_\text{p}$}$\uparrow$ & {SeqCR}$\downarrow$ & {Acc$_\text{p}$}$\uparrow$ & {SeqCR}$\downarrow$ & {Acc$_\text{p}$}$\uparrow$ & {SeqCR}$\downarrow$ \\
\midrule
CoF-7B & 4.1 & 83.1 & 2.6 & 87.8 & 12.7 & 58.0 & 6.8 & 63.5 & 5.8 & 75.8 \\
\rowcolor{rowgray}Qwen2.5-VL-7B & 2.1 & 94.6 & 3.3 & 89.7 & 12.0 & 57.4 & 5.9 & 51.7 & 5.4 & 76.8 \\
PixelReasoner-8B & 2.1 & 95.3 & 1.1 & 92.9 & 12.7 & 60.8 & 2.5 & 69.5 & 4.1 & 81.8 \\
\rowcolor{rowgray}DeepEyes-7B & 0.0 & 100.0 & 1.5 & 94.5 & 8.0 & 72.0 & 1.7 & 94.9 & 2.6 & 90.5 \\
DeepEyesV2-7B & 0.7 & 97.5 & 0.4 & 96.2 & 8.0 & 70.5 & 2.5 & 86.4 & 2.5 & 89.0 \\
\rowcolor{rowgray}Mini-o3-7B & 0.0 & 98.8 & 0.7 & 97.2 & 4.7 & 84.5 & 0.8 & 96.6 & 1.5 & 94.3 \\
\bottomrule
\end{tabular}
}
\end{table*}

\subsection{Evaluation Implementation Details}
\label{sec:eval_impl_app}

Because Collapse Rate compares whether two independent generations produce the \emph{same} answer, the evaluation pipeline runs under a single uniform protocol so that per-model parsing choices do not become a confound for CR. We document the choices below.

\paragraph{API and decoding.} All models are queried through OpenRouter under one configuration: \texttt{temperature=0} (greedy), \texttt{max\_tokens=4096}, no \texttt{top\_p} or seed parameters. Reasoning models that expose an explicit thinking budget (GPT-5-mini and Grok~4.3 in our suite) are run under full reasoning via OpenRouter's \texttt{reasoning=\{"effort":...\}} parameter; all other models use the default low effort. Evaluations were conducted in May~2026.

\paragraph{Prompt.} A single answer-first prompt is used for every image and every model:
\begin{quote}\small\ttfamily
Inspect the image carefully and answer the question.\\
Think internally if needed, but do NOT write your reasoning.\\
Your first line must be: \textless answer\textgreater ...\textless /answer\textgreater\\
Put only the final answer inside the tags. If options are given, answer with the option letter; if it is a number, color, shape, or word, give only that value.\\[2pt]
Question: \{question\}
\end{quote}
The two images of a pair are sent in independent calls under this exact prompt, with no cross-image context.

\paragraph{Answer extraction.} Responses are parsed in a fixed priority order: (1)~\texttt{<answer>...</answer>}; (2)~\texttt{\textbackslash boxed\{...\}}; (3)~``Final Answer:~X'' or ``Answer:~X'' patterns; (4)~``Panel~X'' or ``Panel~(X)'' anywhere in the response; (5)~bracketed \texttt{[X]} (non-LaTeX); (6)~a fallback that returns the last integer found. The inner text from each rule is routed through a small letter-extraction routine that collapses ``Panel~(C)'', ``C.'', and ``C'' into the same canonical token \texttt{C}.

\paragraph{Truncation handling.} If a response hits \texttt{max\_tokens} (\texttt{finish\_reason=length}), we switch to a strict extractor that trusts only an explicit \texttt{<answer>...</answer>} or \texttt{\textbackslash boxed\{...\}} tag. This prevents a cut-off reasoning trace from leaking an intermediate number into the answer slot, which would otherwise contaminate both per-image accuracy and the Collapse Rate.

\paragraph{Normalization for ``same answer''.} The CR test compares $\mathrm{norm}(\hat y_o) = \mathrm{norm}(\hat y_e)$ with $\mathrm{norm}(s) = \texttt{s.strip().upper()}$. An empty extraction on either side is treated as a non-match, so a model that returns \texttt{""} on one side cannot collapse with anything on the other. ``Same answer'' therefore depends only on the canonicalised extracted token, not on the surrounding prose.

\paragraph{Invalid responses.} An invalid response (extractor returns \texttt{""}, typically a truncated reasoning trace with no \texttt{<answer>} tag) is neither correct nor a collapse. It still consumes a pair and reduces the model's competent denominator, so models with elevated truncation rates have a downward-biased CR.

\section{Benchmark and Construction Details}
\label{sec:benchmark_details_app}

\subsection{Relation to Public Benchmarks}
\label{sec:public_benchmarks}

\paragraph{Additional benchmark context.} The main related-work section groups prior benchmarks by role; here we keep a broader map of diagnostic suites that are adjacent but not direct baselines. WHOOPS \citep{bitton2023whoops}, core-knowledge probes \citep{gandhi2024coreknowledge}, intuitive-physics tests \citep{riochet2021intphys}, and chart or olympiad-style mathematical suites \citep{tang2025chartmuseum,gulati2025putnamaxiom,visaidmath2024} stress commonsense, physical, and visual-symbolic reasoning on standalone inputs. They motivate the need for controlled visual reasoning diagnostics, but unlike VisualFLIP they do not pair same-question image edits with deterministic answer flips.

Table~\ref{tab:tech_comparison} situates VisualFLIP relative to existing multimodal reasoning benchmarks. We discuss the key dimensions below.

\paragraph{Paired Image Structure.} Most existing benchmarks (MathVista \citep{lu2024mathvista}, MathVision \citep{wang2024mathvision}, MMMU-Pro \citep{yue2024mmmupro}, V* \citep{wu2024vstar}, VisuLogic \citep{visulogic2025}, MuSLR \citep{muslr2025}) evaluate models on standalone images and measure only outcome correctness, without probing sensitivity to visual changes. BLINK-Twice \citep{blinktwice2025} introduces paired evaluation but with limited coverage (345 images), while VisualSwap \citep{shi2026visualswap} uses image swaps during a single reasoning trajectory to test visual re-examination. VisualFLIP instead provides 1{,}374 images organized as 687 independently queried paired samples, where each pair shares the same question but has deterministically different ground-truth answers, enabling direct measurement of the Collapse Rate.

\paragraph{Image Type and Domain.} Benchmarks vary in their visual domain. MathVista and MMMU-Pro span natural photographs and diagrams; VisuLogic focuses on abstract IQ-test-style images. VisualFLIP targets synthetic and diagrammatic images, where logical reasoning should be precise and unambiguous. This design reduces confounders from natural-image variability, although it does not eliminate all perceptual or editing-related difficulty.

\paragraph{Visual Necessity Mechanism.} Most benchmarks either lack a visual-necessity check (MathVista, VisuLogic) or use coarse heuristics such as text-only baselines (MMMU-Pro, MathVision). BLINK-Twice employs model-failure filtering, selecting samples where models fail despite humans succeeding. VisualFLIP uses a KL-based whole-image masking filter for real-image candidates (Section~\ref{sec:visual_necessity_filter}). The criterion reduces text-answerable cases but does not provide a mathematical guarantee of internal visual grounding.

\paragraph{Evaluation Metrics.} Standard benchmarks report accuracy or its variants such as $\Delta\text{Acc}$. MVI-Bench \citep{chen2025mvibench} further introduces MVI-Sensitivity for robustness to misleading visual inputs, and VisualSwap reports probe accuracy drops under image replacement. VisualFLIP instead reports the \textbf{Collapse Rate} (CR; Eq.~\ref{eq:collapse_rate}), the symmetric, competence-conditioned fraction of pairs on which the model returns the same answer to both images despite the ground-truth flip. This makes the paired perturbation itself the unit of diagnosis and removes the dependence on which side is designated ``original.''

\paragraph{Editing Methodology.} Among benchmarks that involve visual interventions, approaches differ substantially. MVI-Bench studies misleading visual inputs across concept, attribute, and relationship levels; VisualSwap replaces an image during generation with a visually similar but semantically different one. VisualFLIP employs guided symbolic editing: we first extract a structured symbolic representation of task-relevant visual elements, then generate precise editing instructions targeting specific premises in the reasoning chain. The perturbations are minimal, semantically meaningful, and deterministically flip the ground-truth answer.

\begin{table*}[!t]
    \centering
    \scriptsize
    \caption{Positioning \textbf{VisualFLIP} relative to current multimodal reasoning and visual-intervention benchmarks.}
    \label{tab:tech_comparison}
    \resizebox{\textwidth}{!}{%
    \begin{tabular}{@{}lccccclp{1.5cm}@{}}
        \toprule
        \textbf{Benchmark} & \textbf{Year} & \textbf{Size ($N$)} & \textbf{Pairs} & \textbf{Image Type} & \textbf{Editing} & \textbf{Necessity (Filtering)} & \textbf{Metric} \\
        \midrule
        MathVista~\citep{lu2024mathvista} & 2024 & 6,141 & $\times$ & Natural, Synthetic, Diagrammatic & $\times$ & None & $Acc$ \\
        MathVision~\citep{wang2024mathvision} & 2024 & 3,040 & $\times$ & Diagrammatic, Synthetic & $\times$ & Text-only Filtering & $Acc$ \\
        MMMU-Pro~\citep{yue2024mmmupro} & 2024 & 2,115 & $\times$ & Natural, Diagrammatic & $\times$ & Text-only Filtering & $\Delta Acc$ \\
        V*~\citep{wu2024vstar} & 2024 & 192 & $\times$ & Natural & $\times$ & Visual Search Pruning & $Acc$ \\
        \midrule
        BLINK-Twice~\citep{blinktwice2025} & 2025 & 345 & \checkmark & Natural & \checkmark & Model-Failure Filtering & Consist., CoT \\
        VisuLogic~\citep{visulogic2025} & 2025 & 1,000 & $\times$ & Abstract, Synthetic & $\times$ & Caption-blind Validation & $Acc$ \\
        MuSLR~\citep{muslr2025} & 2025 & 1,093 & $\times$ & Multi-domain & $\times$ & Logic-Premise Grounding & $Acc, \text{CoT}$ \\
        MVI-Bench~\citep{chen2025mvibench} & 2025 & 1,248 & $\times$ & Natural, Diagrammatic & \checkmark & Expert Annotation & Acc, MVI-S \\
        VisualSwap~\citep{shi2026visualswap} & 2026 & 800 pairs & \checkmark & Math, Diagrammatic & \checkmark & Similar-pair Curation & Probe Acc, $\Delta$ \\
        \midrule
        \textbf{VisualFLIP (Ours)} & 2026 & 1,374 & \checkmark & Synthetic, Diagrammatic & \checkmark  & KL visual-necessity filter & Acc$_\text{p}$, CR \\
        \bottomrule
    \end{tabular}
    }
\end{table*}


\subsubsection{Real-Image Visual-Necessity Filter}
\label{sec:visual_necessity_filter}

For the real-image source, we first filter candidate MathVision questions to reduce cases that can be answered from text alone. Following the information-bottleneck motivation of retaining task-relevant evidence while suppressing text-answerable shortcuts \citep{tishby2015information}, we compare the policy's option-token distribution under the original image and under a whole-image noise replacement $v_{\text{mask}}$:
\begin{equation}
\label{eq:stage1_kl}
\mathcal{D}_{\text{KL}}\bigl(P(y\mid v_o, q) \,\|\, P(y\mid v_{\text{mask}}, q)\bigr).
\end{equation}
Closed-source APIs lack sequence-level probabilities, so $y$ ranges over the multiple-choice option tokens and per-option log-probabilities are read from the API. Samples whose KL does not exceed $0.1$ are dropped as likely text-answerable. We use a whole-image mask at this stage, before symbolic evidence extraction, to avoid a circular dependency: task-critical boxes identified later are used only in the appendix GMRL study, not for benchmark filtering.

\subsection{VisualFLIP Comprehensive Examples}
\label{sec:vlaze_examples}

\begin{figure*}[!t]
    \centering
    \includegraphics[width=\textwidth]{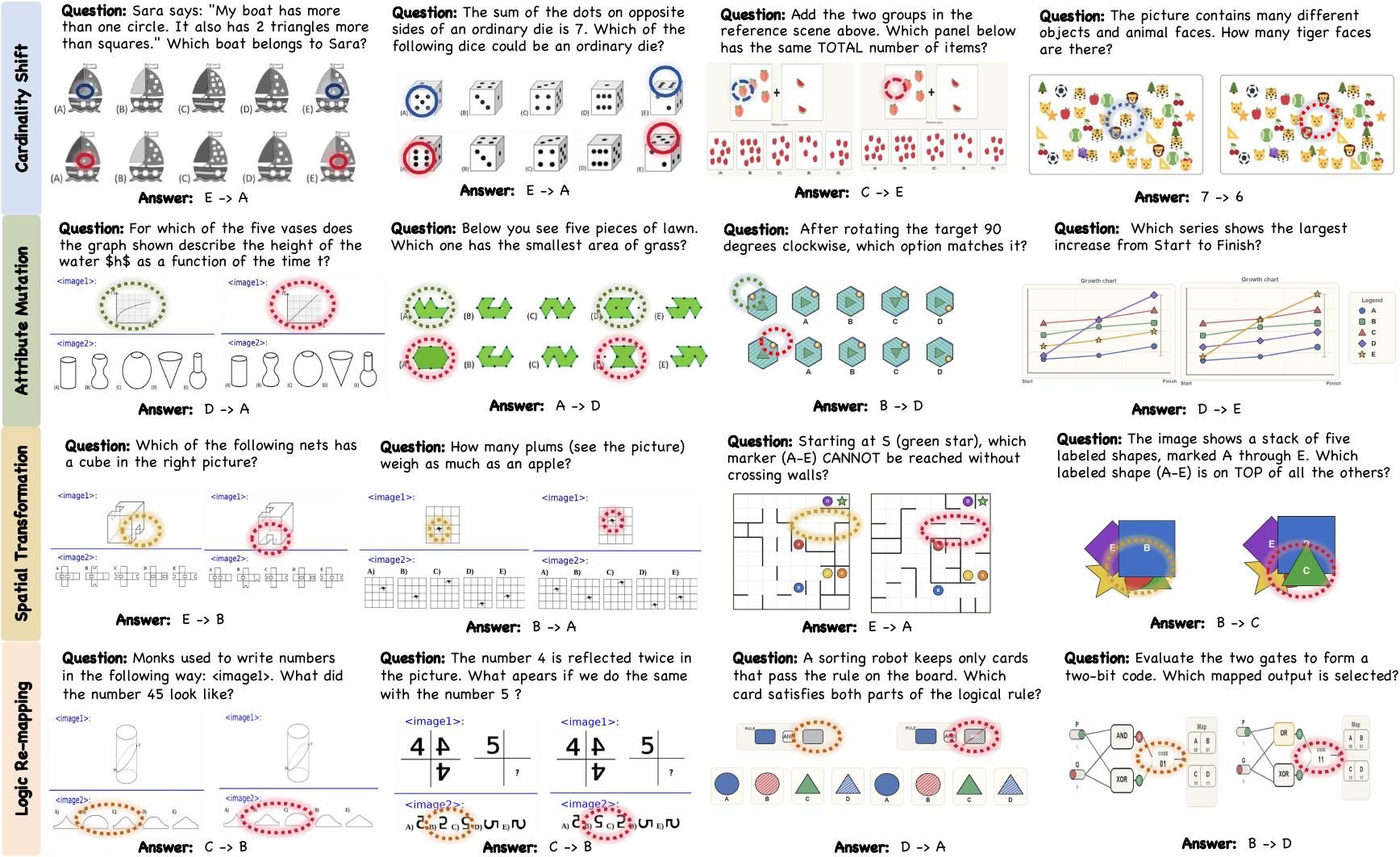}
    \caption[VisualFLIP comprehensive examples.]{\textbf{VisualFLIP Comprehensive Examples.} Representative samples across all four perturbation categories (Cardinality, Attribute, Spatial, Logic). Each row shows an original-edited image pair with the corresponding ground-truth flip. A model whose answer is grounded in the image must update when the visual evidence changes; a model that collapses maintains the original answer despite the change.}
    \label{fig:vlazy_full}
\end{figure*}

Figure~\ref{fig:vlazy_full} presents the complete VisualFLIP evaluation protocol with representative examples from each of the four categories. The benchmark creates minimal visual perturbations that deterministically change the ground-truth answer, exposing whether a model updates its prediction when visual evidence changes.

\paragraph{Cardinality Examples.} The first row demonstrates counting-based reasoning. In the original image, a specific number of objects---shapes on a labeled boat, dots on a die, items aggregated across panels, or tiger faces in a cluttered scene---determines the correct answer. The edited version modifies the count, which should cause a visually grounded model to update its numerical response. Models that collapse maintain the original answer despite the visual evidence of quantity change.

\paragraph{Attribute Examples.} The second row showcases attribute-mutation scenarios where labels, colors, or numerical values are altered. These perturbations target the fundamental premises used in logical deduction. Modifying the slope of a height-time graph, the visible area of a lawn piece, or the visual attributes of a rotation-matching option should shift the model's selection. Models that collapse retain answers derived from the original values.

\paragraph{Spatial Examples.} The third row presents spatial transformations involving positional relationships, trajectories, or geometric configurations. These examples test whether models anchor their spatial reasoning to actual visual coordinates or rely on memorized spatial templates. Perturbations include reconfiguring a 3D cube unfolding, altering a maze path so a target marker becomes reachable, swapping items in a stacked layout, or moving an object within a counting grid.

\paragraph{Logic Examples.} The fourth row illustrates logic re-mapping, where underlying rules or constraints are modified: reflecting a symbol about a different axis, changing the rule a sorting robot enforces, swapping logic gates in a two-bit circuit, or remapping a numeral system. These perturbations require the model to re-reason from new visual constraints.

\subsection{Data Generation Prompts}
\label{sec:prompts}

This section reports the prompts used for the real-image construction route after visual-necessity filtering. Synthetic pairs are generated from programmatic templates and do not use these prompts. We include the core prompt templates rather than the full engineering variants, since the essential contract is the structured extraction of task-critical evidence and the generation of minimal answer-flipping perturbations.

\definecolor{promptbg}{RGB}{255, 255, 255}
\definecolor{promptframe}{RGB}{200, 200, 200}
\definecolor{promptheader}{RGB}{230, 230, 230}

\newtcblisting{promptbox}[1][]{
  enhanced,
  breakable,
  colback=promptbg,
  colframe=promptframe,
  boxrule=0.5pt,
  left=6pt,
  right=6pt,
  top=4pt,
  bottom=4pt,
  toptitle=3pt,
  bottomtitle=3pt,
  colbacktitle=promptheader,
  coltitle=black,
  fonttitle=\sffamily\bfseries\small,
  title={#1},
  listing only,
  listing options={
    basicstyle=\ttfamily\small,
    breaklines=true,
    breakatwhitespace=false,
    columns=fullflexible,
    keepspaces=true,
    aboveskip=0pt,
    belowskip=0pt,
  },
}

\subsubsection{Stage 2: Symbolic Quantization Prompt}

This prompt extracts the task-relevant objects, attributes, and condition-evidence alignment used to derive a controlled perturbation.

\begin{promptbox}[Symbolic Quantization Prompt]
You will analyze a visual reasoning problem. Do not output the final answer.
Only report verifiable visual observations needed for the problem.
If an observation is uncertain, mark it as UNSURE rather than guessing.

[Output Structure (Must Strictly Follow)]

(0) Problem Restatement (1 sentence, objective)

(1) Task Type
- Choose one or more: Cardinality / Attribute / Spatial / Logic.
- Briefly cite the trigger words or visual cues.

(2) Condition List
- Break the problem into C1, C2, ... .
- Each condition must be visually checkable.

(3) Visual Observation Results
- Enumerate only objects relevant to the conditions.
- For each object, provide an ID, location, and key attributes.
- Include counts when the question requires counting.

(4) Condition-Evidence Alignment
- For each condition Ck, list the supporting object IDs.
- Do not conclude the final answer.

(5) Next Step Reasoning Plan (1-3 sentences)
- Explain how the conditions would be combined to obtain the answer.

Problem: \{problem\}
\end{promptbox}

\subsubsection{Stage 3 (a): Counterfactual Editing Plan Generation Prompt}

This prompt turns the symbolic map into a minimal counterfactual perturbation plan whose new answer is deterministic.

\begin{promptbox}[Counterfactual Editing Plan Prompt]
You are now executing the [Counterfactual Image Editing Plan] task.

[Inputs]
- Original image $v_o$
- Question $q$ and original ground-truth answer $y_o$
- Symbolic map $\mathcal{S}$ from Stage 2

[Output Structure (Must Strictly Follow)]
(0) Original answer: state $y_o$.
(1) Target perturbation category (one of: Cardinality / Attribute / Spatial / Logic).
    Cite the condition(s) $C_k$ that justify the choice.
(2) Target object: identify which $O_i$ in $\mathcal{S}$ to modify.
(3) Specific modification action: describe the attribute or relation to change.
(4) New ground-truth answer: derive $y_e \neq y_o$ from the modified evidence.
(5) Freeze list: list visual content that should remain unchanged.
(6) Minimal-change justification: explain why this is the smallest answer-flipping change.

Constraints:
- Do not output the edited image; only produce the plan.
- The new answer $y_e$ must be deterministic; ambiguous flips are forbidden.
- Reject the sample (output "REJECT: <reason>") if no minimal flip exists.

Problem: \{problem\}
Symbolic map: \{symbolic\_map\}
\end{promptbox}

The output is then compressed into a direct image-editing instruction.

\subsubsection{Stage 3 (b): Editing Instruction Compression Prompt}

This prompt converts the perturbation plan into a concise instruction for the image-editing model.

\begin{promptbox}[Editing Instruction Compression Prompt]
You are now executing the [Image Editing Instruction Compression] task.

The input will be a "counterfactual image editing plan" that already contains:
- Original answer and new answer
- Target region or sub-panel, if applicable
- Objects that need to be modified
- Specific modification actions and constraints

Your task:
- Compress the plan into one sentence, or at most two sentences.
- State only what visual changes to make and what should remain unchanged.
- Do not mention answers, condition numbers, or reasoning.
- Use imperative, action-oriented language.

Output Requirements:
- Only output the compressed editing instruction text.
- Do not add titles, numbering, or explanatory notes.
\end{promptbox}

\subsubsection{Stage 4: Image Editing Generation Prompt}

This prompt is used by the image-editing model to realize the compressed perturbation instruction.

\begin{promptbox}[Image Editing Generation Prompt]
You are an expert image editor. Your task is to modify the provided image according to the editing instruction below.

[Editing Instruction]
{instruction}

[Editing Guidelines]
1. Precision: Only modify the elements explicitly mentioned in the instruction.
2. Preservation: Keep all other visual elements visually unchanged.
3. Consistency: Maintain the original style, color palette, and image quality.
4. Realism: Ensure the edited region blends naturally with the surrounding areas.
5. Minimal Change: Apply the smallest modification necessary to achieve the instruction goal.

[Quality Requirements]
- The edited image must be visually seamless and consistent with the original style.
- No visible artifacts, blur, or inconsistencies at edit boundaries.
- Object proportions and perspectives must remain physically plausible.
- Lighting and shadows must be consistent with the scene.

[Output]
Generate the edited image with the specified modifications applied.
\end{promptbox}

\section{Exploratory Mitigation: Grounded Masking Reinforcement Learning}
\label{sec:gmrl_appendix}

\begin{figure}[!h]
  \centering
  \includegraphics[width=\columnwidth]{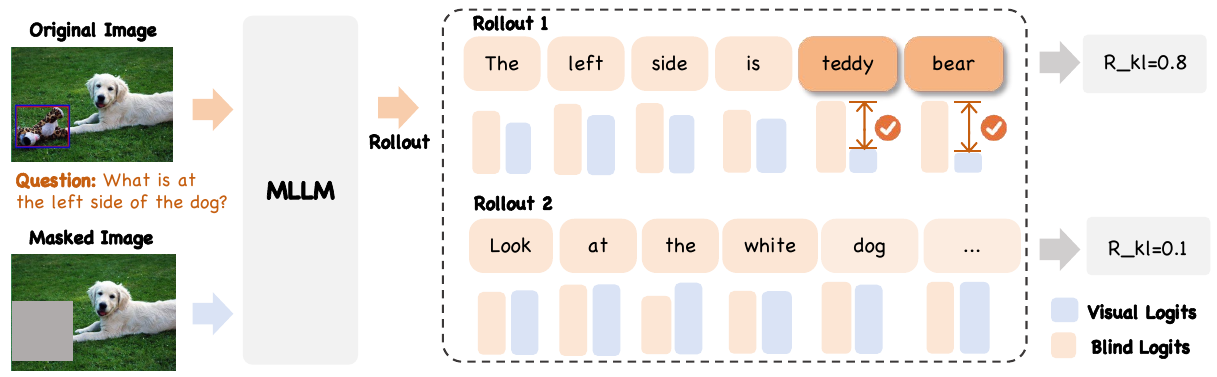}
  \caption[GMRL overview.]{\textbf{GMRL overview.} The visual-necessity reward is the KL divergence between output distributions on the original and locally masked images for the same rollout. Rollouts that attend to the critical region exhibit higher divergence and receive higher reward.}
  \label{fig:method_overview}
\end{figure}

As an exploratory direction, we sketch a training-side reward that is intended to encourage the policy to depend on task-critical visual evidence. \textbf{Grounded Masking Reinforcement Learning (GMRL)} extends GRPO \citep{shao2024deepseekmath} with an auxiliary visual-necessity term: for each rollout we measure how much the policy's token-level distributions change when only the task-critical bounding box is locally masked, and reward rollouts whose answer depended on that region (Figure~\ref{fig:method_overview}).

Let $M_B$ be the binary mask of the task-critical bounding box, and let $v_{\text{mask}} = (1-M_B)\odot v + M_B \odot (0.5\,v + 0.5\,\epsilon)$ with $\epsilon \sim \mathcal{U}[0,1]^{H\times W\times 3}$ be the local perturbation that replaces pixels inside $M_B$ while preserving the rest of the image \citep{chen2025revisiting}. For a rollout $y = (y_1, \dots, y_T)$ generated under the original image, the visual-necessity score is the token-summed KL divergence
\begin{equation}
\label{eq:vn_metric}
\begin{split}
\mathcal{D}_{\text{VN}}(y)
= \sum_{t=1}^{T} D_{\text{KL}}\!\Big(
&\pi_\theta(\cdot \mid y_{<t}, v, q) \,\Big\|\\
&\pi_\theta(\cdot \mid y_{<t}, v_{\text{mask}}, q)\Big).
\end{split}
\end{equation}
We rank-transform $\{\mathcal{D}_{\text{VN}}^{(i)}\}$ within each rollout group and add the bounded auxiliary reward
\begin{equation}
\label{eq:rvn}
\begin{aligned}
\mathcal{R}_{\text{VN}}^{(i)}
&= \mathcal{R}_{\min} + \frac{\rho_i}{G-1}\,(\mathcal{R}_{\max} - \mathcal{R}_{\min}),\\
\mathcal{R}_{\max} &= -\mathcal{R}_{\min} = 0.10,
\end{aligned}
\end{equation}
to the GRPO accuracy reward. Three design choices together limit reward-magnitude hacking: the bounded magnitude cannot flip the ordering of correct versus incorrect rollouts in the group advantage and only reweights rollouts within the same correctness outcome; the rank transform reduces sensitivity to absolute KL scale, rollout length, and prompt difficulty within each group; and the local mask keeps $\pi_\theta(\cdot \mid v_{\text{mask}}, q)$ on the manifold of visual-conditional distributions rather than collapsing it to a question-only prior. These controls limit but do not eliminate the room for the policy to inflate $\mathcal{D}_{\text{VN}}$ by adopting unrelated high-divergence tokens under much longer training.

\paragraph{Setup.} We train Qwen2.5-VL-7B \citep{qwen2025qwen25vl} using $10{,}000$ samples from REF-L4, which supplies the task-critical bounding boxes used for the local mask. Training uses KL coefficient $\beta=0.3$, batch size $128$, $G=8$ rollouts per prompt, $2{,}000$ steps on $8\times$NVIDIA H200 (128GB) GPUs, and a single seed; multi-seed evaluation on additional backbones, including frontier models, is left to future work.

\paragraph{Effectiveness.}
On Qwen2.5-VL-7B, the visual-necessity reward improves both metrics over accuracy-only GRPO on three of the four categories 
(Table~\ref{tab:gmrl_visualflip}). The largest CR reduction is on 
Attribute ($50.4\%\!\to\!45.3\%$), whose answers hinge on a single 
localized value that aligns directly with the REF-L4 bounding-box 
mask. The exception is Logic, on which GMRL slightly underperforms 
GRPO: 
Logic perturbations modify a visible rule (a mirror axis, a sorting 
key, a symbol mapping) rather than a localized object, so an 
object-level box does not cover the load-bearing evidence and the 
masking reward provides little useful gradient.

\begin{table*}[!t]
\centering
\renewcommand{\arraystretch}{1.18}
\setlength{\tabcolsep}{5pt}
\scriptsize
\caption[GMRL on VisualFLIP.]{\textbf{Performance of GMRL on VisualFLIP.} We report Qwen2.5-VL-7B pair accuracy (Acc$_\text{p}$) and Collapse Rate (CR) across perturbation categories. For +GRPO and +GMRL, Avg is the unweighted mean across categories.}
\label{tab:gmrl_visualflip}
\resizebox{\textwidth}{!}{%
\begin{tabular}{llcccccccccc}
\toprule
\multirow{2}{*}{\textbf{Backbone}} & \multirow{2}{*}{\textbf{Method}} & \multicolumn{2}{c}{\textbf{Cardinality}} & \multicolumn{2}{c}{\textbf{Attribute}} & \multicolumn{2}{c}{\textbf{Spatial}} & \multicolumn{2}{c}{\textbf{Logic}} & \multicolumn{2}{c}{\textbf{Avg}} \\
\cmidrule(lr){3-4} \cmidrule(lr){5-6} \cmidrule(lr){7-8} \cmidrule(lr){9-10} \cmidrule(lr){11-12}
& & {Acc$_\text{p}$}$\uparrow$ & {CR}$\downarrow$ & {Acc$_\text{p}$}$\uparrow$ & {CR}$\downarrow$ & {Acc$_\text{p}$}$\uparrow$ & {CR}$\downarrow$ & {Acc$_\text{p}$}$\uparrow$ & {CR}$\downarrow$ & {Acc$_\text{p}$}$\uparrow$ & {CR}$\downarrow$ \\
\midrule
\multirow{3}{*}{Qwen2.5-VL-7B}
 & Base & 3.4 & 45.2 & 12.8 & 61.2 & 6.7 & 53.6 & 11.0 & 35.2 & 9.2 & 53.0 \\
 & + GRPO & 6.7 & 43.2 & 16.4 & 50.4 & 10.2 & 50.3 & 15.4 & 32.6 & 12.2 & 44.1 \\
 & \textbf{+ GMRL (ours)} & 10.2 & 42.2 & 17.8 & 45.3 & 12.2 & 49.8 & 14.4 & 32.9 & 13.7 & 42.6 \\
\bottomrule
\end{tabular}
}
\end{table*}

\subsection{Sampling-Time Analysis}

\begin{figure}[!t]
  \centering
  \includegraphics[width=\columnwidth]{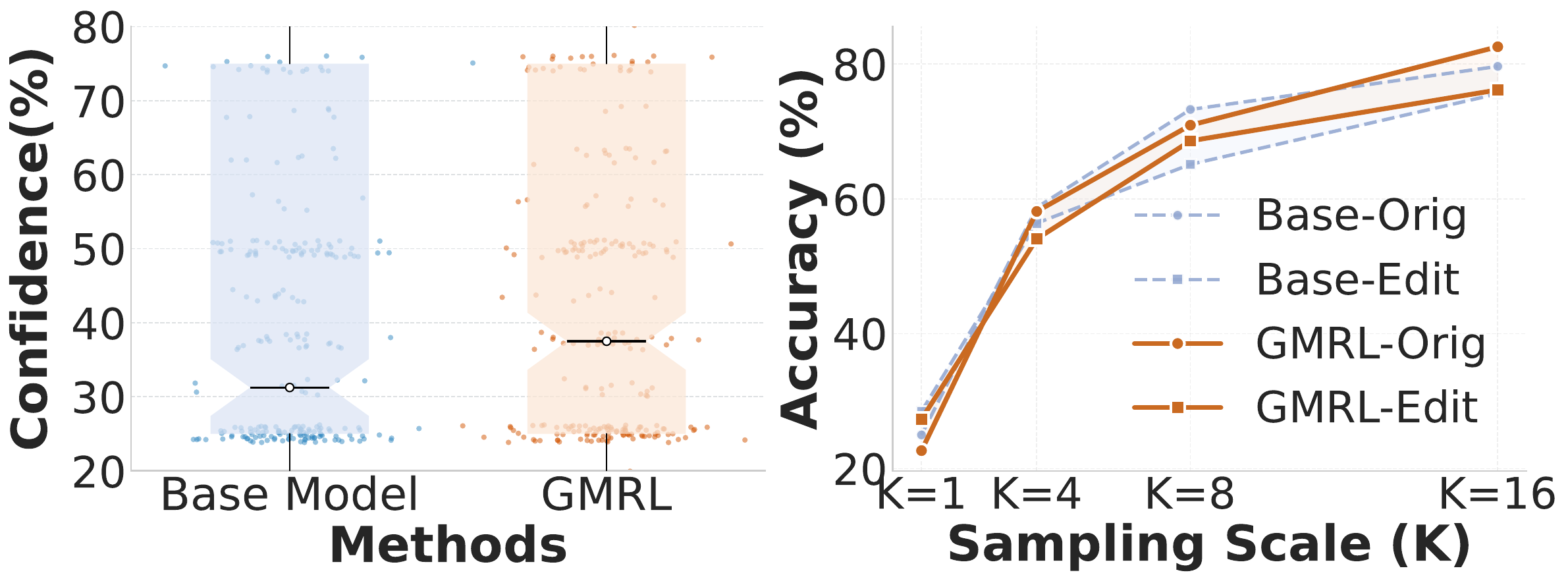}
  \caption[Sampling-time analysis.]{\textbf{Sampling-time analysis on VisualFLIP} (Qwen2.5-VL-7B). \textit{Left:} Pass@$K$ distribution at $K{=}16$. \textit{Right:} Pass@$K$ across budgets; GMRL gains grow with larger $K$. Y-axis reports Pass@$K$.}
  \label{fig:combined_analysis}
\end{figure}

Figure~\ref{fig:combined_analysis} (right) shows how the sampling budget $K$ affects performance on VisualFLIP with the Qwen2.5-VL-7B backbone. The curves exhibit a Pass@$K$ crossover: GRPO leads at $K=1$, where greedy decoding favors the more sharply peaked accuracy-only policy, but GMRL surpasses it as $K$ grows and reaches $82.56\%$ at $K{=}16$ compared to $79.65\%$ for GRPO. We interpret this as evidence that GMRL produces a more diverse set of vision-attentive rollouts that are recoverable under larger sampling budgets, rather than as a broad scaling-law claim, since the analysis is restricted to a single backbone and a single dataset. The left panel shows the corresponding distribution of correct-path density across prompts: GMRL shifts the distribution toward higher values, so the gain at large $K$ is not driven by a few outlier prompts.

\paragraph{Future work.} The current mitigation study is intentionally small: it uses one backbone, one seed, and category-level pair metrics rather than full model-suite evaluation. Future work should rerun GMRL across multiple backbones and seeds, replace REF-L4 boxes with automatically discovered task-critical regions, and separate reward-scale, mask-quality, and sampling-budget effects.




\end{document}